\pdfoutput=1

\documentclass[11pt]{article}

\usepackage[]{EMNLP2022}

\usepackage{times}
\usepackage{latexsym}
\usepackage{graphicx}
\usepackage{algorithm}
\usepackage{algpseudocode}
\usepackage[T1]{fontenc}

\usepackage[utf8]{inputenc}

\usepackage{microtype}

\usepackage{inconsolata}

\usepackage{multirow}
\usepackage{float}
\usepackage[subtle]{savetrees}

\graphicspath{ {images/} }

\title{An Efficient Active Learning Pipeline for Legal Text Classification}

\author{Sepideh Mamooler \and Rémi Lebret \and Stephane Massonnet \and Karl Aberer  \\
        School of Computer and Communication Sciences, EPFL, Switzerland}

\begin{document}
\maketitle
\begin{abstract}
Active Learning (AL) is a powerful tool for learning with less labeled data, in particular, for specialized domains, like legal documents, where unlabeled data is abundant, but the annotation requires domain expertise and is thus expensive. Recent works have shown the effectiveness of AL strategies for pre-trained language models. However, most AL strategies require a set of labeled samples to start with, which is expensive to acquire. In addition, pre-trained language models have been shown unstable during fine-tuning with small datasets, and their embeddings are not semantically meaningful. In this work, we propose a pipeline for effectively using active learning with pre-trained language models in the legal domain. To this end, we leverage the available \textit{unlabeled} data in three phases. First, we continue pre-training the model to adapt it to the downstream task. Second, we use knowledge distillation to guide the model's embeddings to a semantically meaningful space. Finally, we propose a simple, yet effective, strategy to find the initial set of labeled samples with fewer actions compared to existing methods. Our experiments on Contract-NLI, adapted to the classification task, and LEDGAR benchmarks show that our approach outperforms standard AL strategies, and is more efficient. Furthermore, our pipeline reaches comparable results to the fully-supervised approach with a small performance gap, and dramatically reduced annotation cost. Code and the adapted data will be made available.
\end{abstract}
  
\section{Introduction}

With the advent of pre-trained transformer-based language models (\citealp{Devlin2019BERTPO,Liu2019RoBERTaAR,He2021DeBERTaDB}), training models from scratch has been outperformed by fine-tuning pre-trained language models for several tasks in natural language processing, including text classification \citep{Howard2018UniversalLM}. However, fine-tuning these models still needs large labeled datasets to perform well on the downstream task \citep{Dodge2020FineTuningPL,Zhang2021RevisitingFB,Mosbach2021OnTS}. Collecting a large annotated dataset is a highly expensive and time-consuming process in specialized domains, where annotation can only be performed by the domain experts, such as the legal domain \citep{Hendrycks2021CUADAE}.

Active Learning (AL) has been proved effective for data-efficient fine-tuning of pre-trained language models in non-specialized domains like news, emotions, and movies \citep{EinDor2020ActiveLF,Margatina2022OnTI}. In addition, \citet{Margatina2022OnTI} have shown that the unlabeled data can be used to adapt the pre-trained language model to the downstream task, thereby improving the active learning performance with no extra annotation cost. On the specialized domains, \citet{Chhatwal2017EmpiricalEO} have evaluated multiple AL strategies in the legal domain before the emergence of pre-trained language models. Nevertheless, to the best of our knowledge, the effectiveness of active learning in fine-tuning pre-trained language models in the legal domain has been poorly studied. 

In this work, we focus on efficient legal text classification with RoBERTa \citep{Liu2019RoBERTaAR} by leveraging existing AL strategies. We identify two challenges in deploying AL strategies in the legal domain; First, legal texts contain a specialized vocabulary that is not common in other domains, including the ones on which pre-trained language models are trained. Second, the annotation of legal texts is highly expensive and time-consuming due to the necessity of specialized training for understanding these texts. For example, \citet{Hendrycks2021CUADAE} reported a cost of over $\$2$ million for the annotation of the Contract Understanding Atticus Dataset (CUAD) consisting of around $500$ contracts. 

To account for the specialized vocabulary, inspired by \citeposs{Margatina2022OnTI} work, we leverage the available \textit{unlabeled} data to adapt the pre-trained language model to the downstream task. In addition, considering the limitations of pre-trained language models like BERT and RoBERTa in capturing semantics \citep{Reimers2019SentenceBERTSE}, we use knowledge distillation to further improve the task-adapted model by mapping its embedding space to a semantically meaningful space. Our experiments demonstrate that AL strategies can benefit from semantically meaningful embeddings.

Concerning the cost and time constraints, we focus on the fact that many AL strategies \citep{Lewis1994ASA,Gal2016DropoutAA,Gissin2019DiscriminativeAL} require an annotated set of $N$ positive and negative samples to start with. In practice, acquiring this set is expensive for large and skewed datasets. We propose a strategy to make the first iteration more efficient by clustering the unlabeled samples and limiting the pull of candidates to the cluster medoids. Our experiments demonstrate we can achieve comparable results with the standard initial sampling approach with up to $63\%$, and $25\%$ fewer actions on the skewed Contract-NLI \citep{Koreeda2021ContractNLIAD}, and balanced LEDGAR benchmarks \citep{Tuggener2020LEDGARAL} respectively. 

Our contributions can be summarized as follows:

\begin{itemize}
    \item[1.] We design an efficient and effective active learning pipeline for legal text classification by leveraging the available unlabeled data using task-adaptation and knowledge distillation, which obtains comparable performance to fully-supervised fine-tuning with considerably reduced annotation effort.
    \item[2.] We propose a strategy to reduce the number of actions in the first iteration of active learning by clustering the unlabeled data, and collecting the samples from cluster medoids, further increasing the efficiency of our approach.
    \item[3.] We evaluate our approach over Contract-NLI and LEDGAR benchmarks. Our results illustrate an increase of $0.3346$, and $0.1658$ in the best obtained F1-score, compared to standard active learning strategies, for Contract-NLI and LEDGAR respectively.
    
\end{itemize}

\section{Related Work}

\paragraph{Active learning with pre-trained language models}
Multiple works have studied active learning for pre-trained language models like BERT. \citet{EinDor2020ActiveLF} have evaluated various AL strategies for fine-tuning BERT for text classification, and showed that AL can boost BERT's performance especially for skewed datasets. However, they do not leverage the available unlabeled data to adapt the pre-trained language model to the task at hand, and only focus on non-specialized domains like news and sentiment analysis that do not require experts' knowledge.

\citet{Gururangan2020DontSP} have shown that task-adaptive pre-training using the available unlabeled data leads to performance gain when using pre-trained language models like BERT. Following this observation, \citet{Margatina2022OnTI} demonstrated the importance of task-adaptation for active learning for non-specialized texts like news, movies and sentiment analysis. 

Inspired by these works, we leverage the available unlabeled data to effectively adapt RoBERTa to legal text classification, where the annotation demands experts' knowledge. In addition, we propose an additional step to map the embedding space of the task-adapted RoBERTa to a semantically meaningful space using sentence transformers. 

\paragraph{Sentence transformers} 
\citet{Reimers2019SentenceBERTSE} have shown that the embedding space of off-the-shelf pre-trained language models like BERT \citep{Devlin2019BERTPO} and RoBERTa \citep{Liu2019RoBERTaAR} is not semantically meaningful, and thus, is not suitable for common sentence comparison measures like cosine similarity. To overcome this limitation, they propose sentence transformers, obtained by adding a pooling layer on top of pre-trained language models, and fine-tuning them in a Siamese network architecture with pairs of similar sentences. In this work, we use a RoBERTa-based sentence transformer as a teacher model and distill its knowledge to the task-adapted RoBERTa to produce sentence embeddings that capture the semantics and can be compared using cosine similarity. 

\paragraph{Active learning strategies}
Numerous methods have been proposed to find proper labeling candidates for active learning. Majority of them belong to one or both of two categories: diversity-sampling, and uncertainty-sampling. Diversity-based methods (\citealp{Sener2018ActiveLF,Gissin2019DiscriminativeAL,Wang2017IncorporatingDA}) aim to find labeling candidates that best represent the dataset, whereas uncertainty-based methods (\citealp{Gal2016DropoutAA,Kirsch2019BatchBALDEA,Zhang2021CartographyAL}) target candidates about which the model is uncertain. 
BADGE \citep{Ash2020DeepBA} is a cluster-based AL strategy that belongs to both of these categories. It transforms data into gradient embeddings that encode model confidence and sentence feature at the same time. 
By applying kmeans++ on the gradient embeddings it can find samples that differ both in terms of semantics and predictive uncertainty. ALPS \citep{Yuan2020ColdstartAL} is another cluster-based AL strategy that leverages both uncertainty and diversity using the surprisal embeddings obtained by passing the sentences to the MLM head of the pre-trained language model, and computing the cross entropy loss for a random set of tokens against the target labels. 

Existing AL strategies often require a set of labeled samples to start with, which is expensive to acquire. To overcome this high cost, we propose a clustering-based strategy to reduce the effort required to create the initial set of annotated samples.

\section{Notation and Setting}
In this section, we explain the structure shared between all AL strategies used in this work and fix the notation.

Active learning is an iterative process aiming to obtain a desired performance given an annotation budget. Here, we consider the annotation budget to be the number of actions performed by the annotator. In addition, we assume all annotators are legal experts, and that each annotator assigns perfect labels to text segments. Let $U_0$ and $L_0$ be the starting pool of unlabeled and labeled samples respectively. Initially, $L_0 = \emptyset$. At the first iteration, the annotator labels $N$ sample, $P$ positive and $N-P$ negative, to obtained $L_1$. Then, at each iteration $i$, the model is fine-tuned using $L_i$, and the AL strategy recommends a set of samples $C_i$ for annotation. These samples are labeled and $U_i$ and $L_i$ are updated as $U_{i+1} = U_i \setminus C_i$, and $L_{i+1} = L_i \cup C_i $. The procedure is repeated until the annotation budget is exhausted, or the desired performance is achieved.

We base our work on the Low-Resource Text Classification Framework introduced by \citet{EinDor2020ActiveLF}. 
Following this work, we focus on binary text classification, given a small annotation budget and a potentially imbalanced dataset. This scenario matches common use cases in the legal domain, where the goal is to find phrases that correspond to a specific category, with the lowest possible number of actions, given a pool of unlabeled, imbalanced data. We perform $5$ AL iterations, and assume a more restricted annotation budget compared to \citet{EinDor2020ActiveLF}, allowing only $10$ annotations per iteration. For the first AL iteration, we assume that $5$ positive and $5$ negative samples need to be annotated.

\section{Methodology}

\begin{algorithm*}[ht]
\caption{AL pipeline for text classification}\label{alg:cap}
\hspace*{\algorithmicindent} \textbf{Input:} {unlabeled samples $U_0$, PT RoBERTa, PT Sentence-RoBERTa, AL strategy $\alpha$, \# iterations $T$}
\hspace*{\algorithmicindent} \textbf{Output:} {text classifier CLS RoBERTa, acquired labeled dataset $L_T$}
\begin{algorithmic}
\State $L_0 \gets \emptyset$
\State \textbf{\textit{Phase 1: Task-adaptation with Masked Language Modeling (MLM)}} \State TAPT RoBERTa $\gets$ MLM(PT RoBERTa, $U_0$) 
\State \textbf{\textit{Phase 2: Knowledge distillation}} 
\State DisTAPT RoBERTa $\gets$ Distill(TAPT RoBERTa, PT Sentence-RoBERTa, $U_0$)
\State \textbf{\textit{Phase 3: Initial sampling}} 
\State cluster medoids $\gets$ KMeans(DisTAPT RoBERTa, $U_0$)
\State $L_1 \gets$ Sample(cluster medoids)
\State $U_1 \gets U_0 \setminus L_1$
\State \textbf{\textit{Phase 4: Active learning}} 
\For{$i \gets 1$ to $T$ }
    \State CLS RoBERTa $\gets$ Train(DisTAPT RoBERTa, $L_i$)
    \State $C_i \gets \alpha$(CLS RoBERTa, $U_i$)
    \State $L_{i+1} \gets L_i \cup C_i$
    \State $U_{i+1} \gets U_i \setminus C_i$
\EndFor
\end{algorithmic}
\label{alg:algorithm}
\end{algorithm*}

We propose an efficient active learning pipeline for fine-tuning pre-trained language models for legal text classification. Our approach leverages available unlabeled data in three phases to adapt the pre-trained model to the downstream task (Sec.~\ref{sec:task-adaptation}), guide its embedding space to a semantically meaningful and comparable space (Sec.~\ref{sec:distillation}), and reduce the number of actions required to collect the initial labeled set (Sec.~\ref{sec:sampling}). Finally, it leverages existing AL strategies to efficiently fine-tune a classifier (Sec.~\ref{sec:meth_al}). We now explain each step in detail. An overview of this pipeline can be found in Algorithm~\ref{alg:algorithm}.

\subsection{Task-Adaptation}\label{sec:task-adaptation}

It has been shown that fine-tuning off-the-shelf pre-trained language models with standard approaches is unstable for small training sets \citep{Zhang2021RevisitingFB,Dodge2020FineTuningPL,Mosbach2021OnTS}. As shown by \citet{Margatina2022OnTI}, this can lead to poor performance when fine-tuning pre-trained language models with AL. In addition, existing pre-trained language models are often trained on texts that do not need specialized training to be understood. However, legal texts contain specialized words that are not common in other domains. Thus, task-adaptation is crucial for the effectiveness of active learning in legal text classification. In the first step of our proposed pipeline, we obtain the task-adapted pre-trained (TAPT) RoBERTa by continuing pre-training the model with unlabeled samples for the Masked Language Modeling (MLM) task, as suggested by \citet{Gururangan2020DontSP} and \citet{Margatina2022OnTI}. 
 
\subsection{Knowledge Distillation}\label{sec:distillation}

Previous works \citep{Reimers2019SentenceBERTSE,Li2020OnTS,Su2021WhiteningSR} have shown that, without fine-tuning, the sentence embeddings produced by pre-trained language models poorly capture semantic meaning of sentences, and are not comparable using cosine similarity. To overcome this shortcoming, \citet{Reimers2019SentenceBERTSE} introduced sentence transformers by adding a pooling layer on top of pre-trained transformer-based language models, and training them in a Siamese network architecture with pairs of similar sentences. Compared to out-of-the-box RoBERTa, a RoBERTa-based sentence transformer drives semantically comparable sentence embeddings. 

As we will explain in Sec.~\ref{sec:sampling}, we cluster the normalized sentence embeddings based on their Euclidean distance to efficiently acquire the labeled samples for the initial iteration of AL. The Euclidean distance between normalized embeddings can be driven from their cosine similarity. Hence, sentence embeddings that are comparable with cosine similarity can result in clusters with higher quality. In addition, semantically meaningful sentence embeddings give a better initialization of the \texttt{[CLS]} token, thereby obtaining better classification performance with a smaller training set.

We use a pre-trained RoBERTa-based sentence transformer (PT Sentence-RoBERTa) as a teacher model, and distill its knowledge to the TAPT RoBERTa. The resulting distilled task-adapted pre-trained (DisTAPT) RoBERTa produces semantically meaningful embeddings that are comparable via cosine similarity, and, as shown by our experiments (Sec.~\ref{sec:exp_distillaition}), benefit the classification task.

\subsection{Initial Sampling}\label{sec:sampling}

Many AL strategies (\citealp{Gissin2019DiscriminativeAL,Gal2016DropoutAA}) require an initial set of $N$ labeled samples containing $P$ positive and $N-P$ negative sentences, which is either assumed to be available, or obtained by randomly sampling the entire dataset until the desired number of positive and negative samples are found. This approach is highly expensive for large and skewed datasets. We propose a simple, yet effective, strategy to efficiently acquire the initial labeled set. To this end, we leverage the distilled task-adapted pre-trained RoBERTa to cluster the unlabeled samples using KMeans algorithm \citep{MacQueen1967SomeMF}. The labeled set for the initial iteration is then driven from the cluster medoids. As a result, we shrink the pool of candidates from the entire dataset to the cluster medoids, therefore, reduce the number of actions for obtaining the initial annotated set, while achieving comparable performance with the standard approach for initial sampling.

\subsection{Active Learning}\label{sec:meth_al}
In the last phase, we iteratively fine-tune the DisTAPT RoBERTa for the downstream task. The initial labeled set is used at the first iteration. Then, more samples are labeled in the following rounds using an AL acquisition strategy until the annotation budget is exhausted, or the classifier satisfies the expected performance. 

Our proposed pipeline can be used with existing AL strategies and, as demonstrated by our experiments (Sec.~\ref{sec:exp_distillaition}), consistently outperforms standard AL approaches, regardless of the AL strategy used.

\section{Experimental Setup}\label{sec:setup}

We evaluate our approach against four standard active learning strategies provided in the Low-Resource Text Classification Framework \citep{EinDor2020ActiveLF}:

\begin{itemize}
    \item \textbf{Random} At each iteration, this approach randomly chooses samples for annotation.
    \item \textbf{Hard-Mining} Selects instances that the model is  uncertain about, based on the absolute difference of prediction score and $0.5$. 
    \item \textbf{Perceptron Dropout} \citep{Gal2016DropoutAA} Also selects instances for which the model is least certain. The uncertainty is calculated using Monte Carlo Dropout on $10$ inference cycles. 
    \item \textbf{Discriminative Active Learning (DAL)} \citep{Gissin2019DiscriminativeAL} Deploys a binary classifier to select instances that best represent the entire unlabeled samples. 
\end{itemize}

We consider pre-trained RoBERTa and LEGAL-BERT~\citep{Chalkidis2020LEGALBERTTM} as the baselines. Note that our goal is not to rely on domain-adapted models like LEGAL-BERT since they might not always be available. For example, if the data is in German, we can find a pre-trained RoBERTa in German, but the LEGAL-BERT is pre-trained on English text only.

\subsection{Datasets}\label{sec:datasets}
We evaluate our framework on Contract-NLI \citep{Koreeda2021ContractNLIAD} and LEDGAR \citep{Tuggener2020LEDGARAL} benchmarks. 

Contract-NLI \citep{Koreeda2021ContractNLIAD} is a dataset for document-level natural language inference. It consists of $607$ documents with $77.8$ spans per document on average. Each span is checked against $17$ hypotheses and classified as contradiction, entailment, or not mentioned. In this work, we adapt this dataset to the classification task by considering each hypothesis as a category. If a span is classified as contradiction or entailment for a hypothesis, we label it with the corresponding category. Following this approach, we end up with a classification dataset with $4,371$ train, $614$ development, and $1,188$ test samples within $17$ classes.

LEDGAR \citep{Tuggener2020LEDGARAL} is a text classification benchmark consisting of a corpus of legal provisions in contracts. The entire dataset consists of $846,274$ provisions and $12,608$ labels. We only consider a subset of this dataset that corresponds to provisions with labels that appeared at least $10,000$ times in the corpus, resulting in $44,249$ train, $7,375$ development, and $12,907$ test samples across $5$ categories. Similar to \citet{Tuggener2020LEDGARAL}, we perform a $70\%-10\%-20\%$ random split to obtain the train, development and test sets. 

The class distributions of both datasets can be found in the appendix (Sec.~\ref{sec:app_class_dist}). Compared to Contract-NLI, LEDGAR has fewer categories, is an order of magnitude bigger, and is more balanced.

\subsection{Implementation Details}

We base our implementation on the Low-Resource Text Classification Framework provided by \citet{EinDor2020ActiveLF}\footnote{\url{https://github.com/IBM/low-resource-text-classification-framework}}, and augment it with the task-adaptation, knowledge distillation, and initial sampling steps. 

As the pre-trained model, we use \texttt{roberta-base}\footnote{\url{https://huggingface.co/roberta-base}} (with $125$M parameters), the RoBERTa \citep{Liu2019RoBERTaAR} language model trained on the union of $5$ datatsets: Book corpus \citep{Zhu2015AligningBA}, English Wikipedia\footnote{\url{https://dumps.wikimedia.org}}, CC-News \citep{Mackenzie2020CCNewsEnAL}, OpenWebText Corpus \citep{Gokaslan2019OpenWeb}, and Stories \citep{Trinh2018ASM}, none of which belong to the legal domain. 

For LEGAL-BERT, we use the \texttt{nlpaueb/legal-bert-base-uncased}\footnote{\url{https://huggingface.co/nlpaueb/legal-bert-base-uncased}} (with $110$M parameters), trained on six datasets containing legal docments across Europe and the US.

For task-adaptation, we continue pre-training RoBERTa for the MLM task using the available unlabeled data. We train for $10$ epochs with batch-size $64$, and the learning rate set to $3\mathrm{e}{-4}$. The task-adapted model has perplexity $4.9706$ for Contract-NLI and $2.1628$ for LEDGAR.

For model distillation, we use \texttt{stsb-roberta-base-v2} (with $125$M parameters), a RoBERTa-based sentence transformer trained on the STS benchmark \citep{Cer2017SemEval2017T1}, as the teacher model, and the task-adapted RoBERTa as the student model. Mean Squared Error (MSE) is used as the loss function. The student model is trained for $10$ epochs, with $10$K warmup steps, $1\mathrm{e}{-4}$ learning rate and no bias correction. The final MSE ($\times 100$) is $6.8607$ for Contract-NLI, and $7.2003$ for LEDGAR.

For clustering the normalized sentence embeddings we use the KMeans implementation by \texttt{scikit-learn}. We cluster the Contract-NLI and LEDGAR sentence embeddings into $437$, and $442$ groups respectively. The number of clusters are chosen based on the dataset size, and the number of categories, and to make initial sampling with cluster medoids manageable for experts. 

In all the active learning experiments, we perform $5$ AL iterations, starting with $10$ initial samples, and increasing the size of the annotated data by $10$ at each iteration. Adam optimizer \citep{Kingma2015AdamAM} is used with learning rate set to $5\mathrm{e}{-5}$. The model is trained for $100$ epochs and early stopping is used with patience set to $10$. To account for randomization, we repeat each experiment three times.

To compare our approach with standard AL methods, we use F1-score as the evaluation metric as it captures both precision and recall and is sensitive to data distribution.

\section{Results and Discussion}

In this section, we provide the results of our experiments and explain them in detail. We start by comparing our approach with and without the initial medoid sampling against standard AL strategies (Sec.~\ref{sec:ours-vs-baseline}). Then, we show the effectiveness of knowledge distillation on top of task-adaptation (Sec.~\ref{sec:exp_distillaition}). In addition, we demonstrate the efficiency of the initial sampling with cluster medoids (Sec.~\ref{sec:exp_initial-sampling}). Finally, we evaluate how well our approach performs for different AL strategies (Sec.~\ref{sec:exp_AL_effect}). 

\begin{figure}[t]
    \centering
    \includegraphics[width=0.5\textwidth]{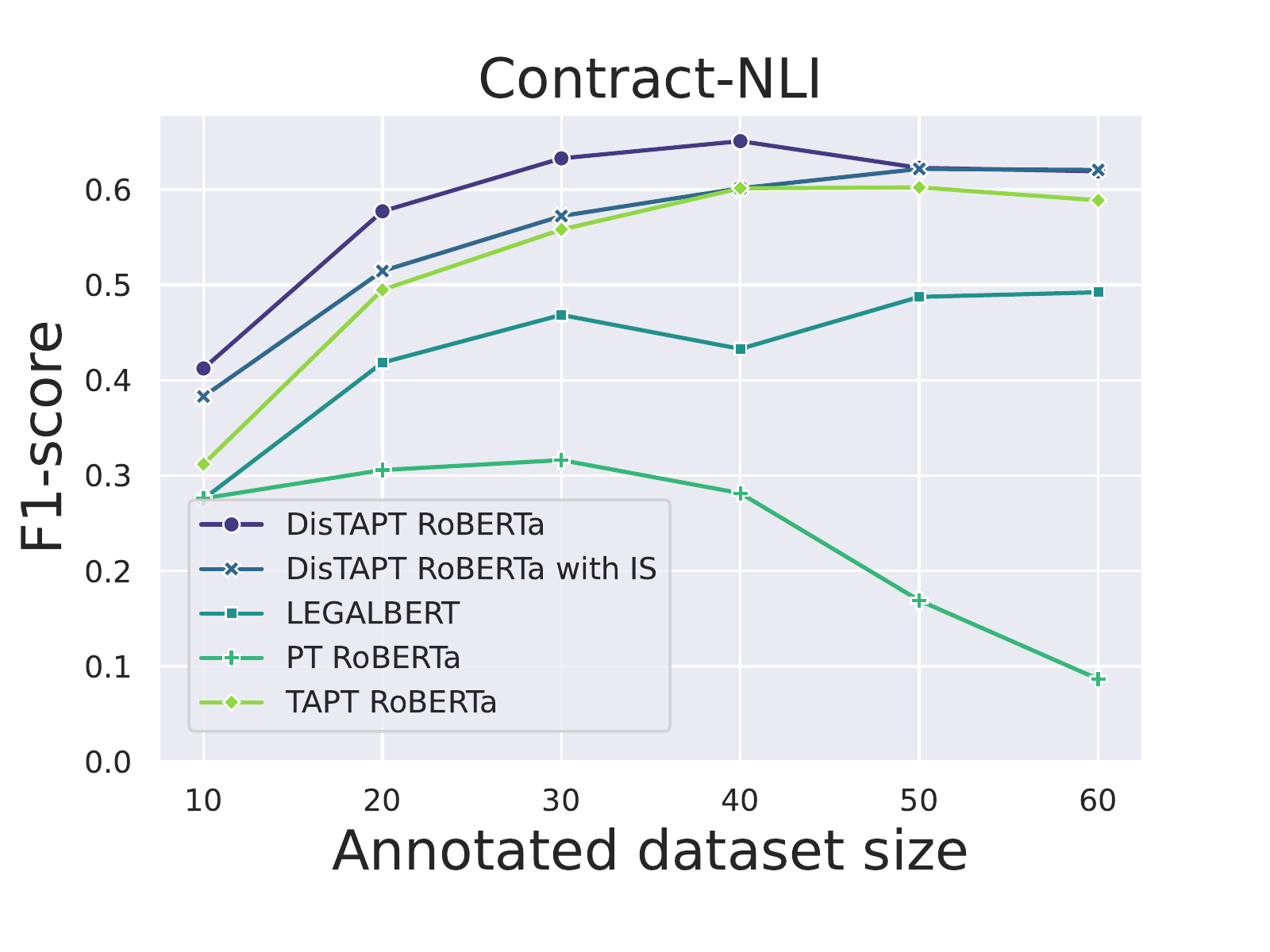}
    \includegraphics[width=0.5\textwidth]{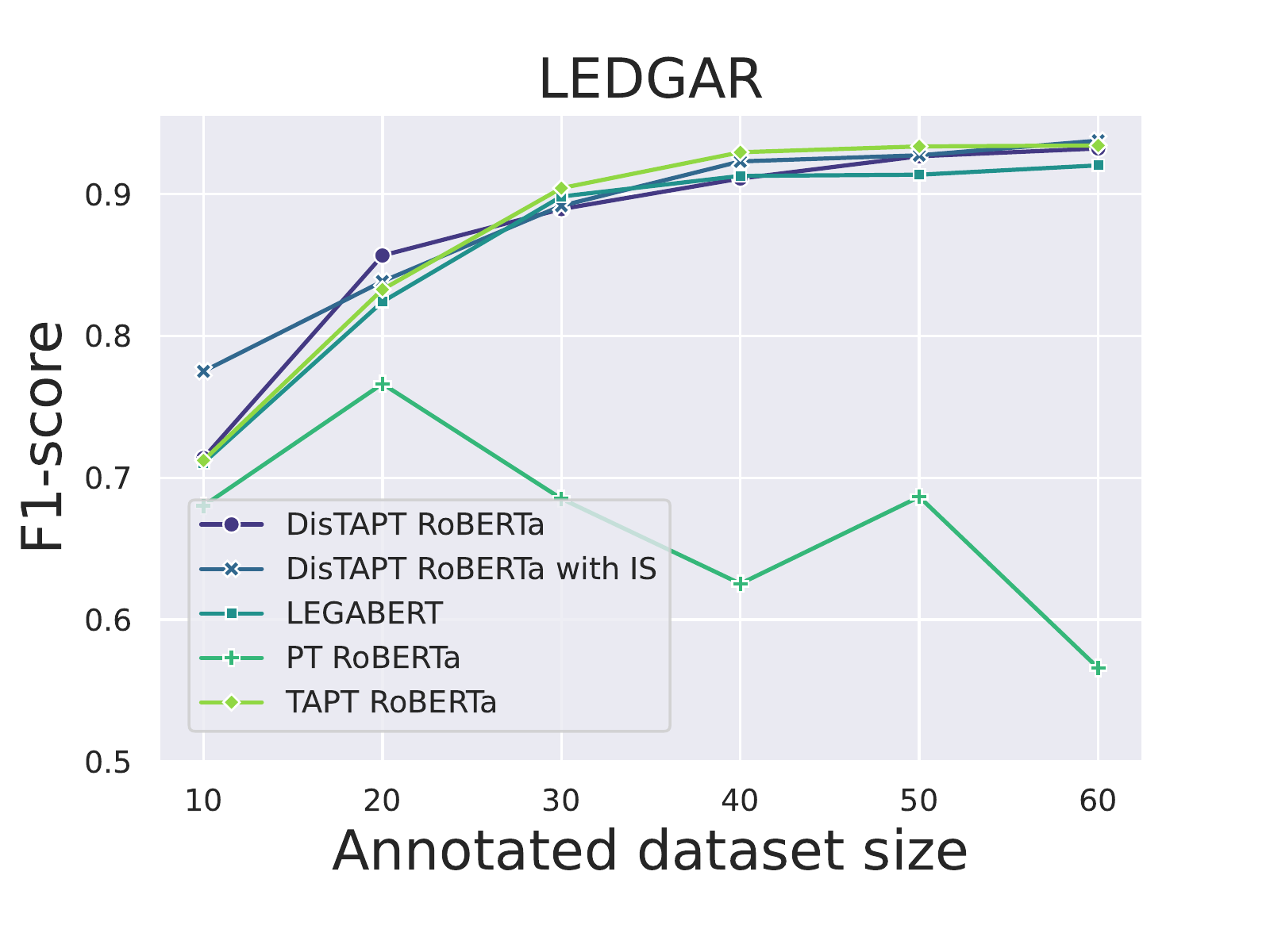}
    \caption{Test F1-score for \textbf{DAL} during AL iterations. The F1-score for the fully supervised fine-tuning is $0.6990$ for Contract-NLI and $0.9538$ for LEDGAR. The figure is best viewed in color.}
    \label{fig:ours-vs-baseline}
\end{figure}

\subsection{Efficient AL Pipeline}\label{sec:ours-vs-baseline}

Figure~\ref{fig:ours-vs-baseline} compares our approach with and without the initial sampling phase (DisTAPT with IS, and DisTAPT) to standard DAL with pre-trained (PT) RoBERTa, LEGALBERT, and TAPT RoBERTa for Contract-NLI and LEDGAR benchmarks. We report the average F1-score over all categories. DAL is chosen due to its better performance, as shown in Figure~\ref{fig:all_ALs}. 
The results for other AL strategies can be found in the appendix (Sec.~\ref{sec:app_tabs}). 

Our experiments show the importance of task-adaptation and knowledge distillation for pre-trained language models prior to fine-tuning with active learning. Figure~\ref{fig:ours-vs-baseline} illustrates that, for the same size of annotated data, our pipeline consistently achieves better performance than standard AL approaches even for LEGAL-BERT.

For the Contract-NLI dataset, the F1-score obtained by fully-supervised fine-tuning (with $4,371$ labeled samples) is $0.6990$ for \texttt{roberta-base} and $0.7152$ for \texttt{legal-bert-base-uncased}. DisTAPT RoBERTa reaches a F1-score as high as $0.6508$ with only $40$ labeled samples. The best F1-score obtained using pre-trained RoBERTa is $0.3162$ with $30$ labeled samples, which is $0.3165$ lower than DisTAPT RoBERTa's F1-score for the same size of annotated data.

For the LEDGAR dataset, the F1-score obtained by the fully-supervised fine-tuning (with $44,249$ labeled samples) is $0.9538$ for \texttt{roberta-base} and $0.9588$ for \texttt{legal-bert-base-uncased}. DisTAPT RoBERTa reaches a very close performance of $0.9321$ F1-score with merely $60$ labeled samples. The highest F1-score that pre-trained RoBERT reaches is $0.7663$ with $20$ annotated samples, which is $0.0904$ lower than DisTAPT's performance with the same size of labeled data.

These results show that, for both datasets, there is only a small performance gap between our approach and the fully-supervised approach, indicating that our AL pipeline dramatically reduces the annotation cost, while achieving comparable performance with the fully-supervised fine-tuning.

In addition, It is observed that standard AL with off-the-shelf pre-trained RoBERTa is unstable. This is aligned with the previous works' observations \citep{Mosbach2021OnTS,Zhang2021RevisitingFB,Dodge2020FineTuningPL}. During fine-tuning, the pre-trained model should perform two tasks: adaptation to the legal domain with the new vocabulary, and classification. By performing task-adaptation and knowledge distillation before fine-tuning, we train the model in a curriculum learning approach, making the model stable even for small training sets.

\subsection{Effect of Knowledge Distillation}\label{sec:exp_distillaition}

To evaluate the effectiveness of knowledge distillation on the quality of obtained clusters, we compare the distribution of the Dunn Index of the clusters before and after knowledge distillation. For both datasets, after knowledge distillation, most of the clusters have higher Dunn Index which indicates that they are more compact and better separated than the clusters before knowledge distillation step. The results are provided in the appendix~\ref{sec:app_knowldge_distill} due to space constraints.

In addition, we evaluate the effect of knowledge distillation on the task-adapted pre-trained RoBERTa, and report the average F1-score over all classes for each dataset. Figure~\ref{fig:ours-vs-baseline} shows that, for both datasets, DisTAPT RoBERTa outperforms TAPT RoBERTa at early iterations of active learning, and as the size of the labeled set increases, the two models' performance converge. This can be explained by the fact that, initially, DisTAPT RoBERTa's embeddings better capture the semantics of sentences, and thus result in better classification performance. As the labeled data grows, TAPT RoBERTa is fine-tuned and can produce semantically meaningful embeddings as well. Hence, for a highly restricted annotation budget, distilling the knowledge of a sentence transformer to the TAPT language model can lead to performance gain.

\subsection{Efficiency of Initial Medoid Sampling} \label{sec:exp_initial-sampling}

It was shown in Figure~\ref{fig:ours-vs-baseline} that DisTAPT with IS obtains comparable performance with DisTAPT without IS. In this section, we evaluate the \textit{efficiency} of the proposed sampling strategy for the initial iteration of AL. 

To this end, we simulate the standard sampling strategy by randomly sampling text segments from the full dataset until $5$ positive and $5$ negative samples are found. The number of iterations is then considered as the number of annotations required to collect the labeled set for the initial AL iteration. Similarly, to simulate our proposed initial sampling, we randomly sample from cluster medoids until $5$ positive and $5$ negative samples are obtained. To account for randomness, we repeat the simulations $1000$ times and report the median and the $90^{th}$ percentile over all runs.  

Table~\ref{tab:exp_medoid} illustrates the results of our simulations for Contract-NLI and LEDGAR. Due to the high number of classes in Contract-NLI, only eight categories of this dataset are presented in this table, and the results for other categories can be found in the appendix (Sec.~\ref{sec:app_medoid}). For each class, in addition to the median and $90^{th}$ percentile over $1000$ runs, the difference in the $90^{th}$ percentile between standard approach and our strategy (in $\%$) is reported as the gain in annotation effort. For example, for the \texttt{Sharing with third-parties} class in Contract-NLI, the $90^{th}$ percentile is $62\%$ less when using medoids for initial sampling, meaning that, with $90\%$ confidence, the annotators perform $62\%$ fewer actions to acquire the initial labeled set using our approach. 

It is observed that, for the skewed Contract-NLI dataset, our proposed initial sampling strategy reduces the number of actions performed by the annotator up to $63\%$. For LEDGAR however, which consists of balanced categories, the highest effort gain in sampling from cluster medoids is $25\%$. There are also few cases where using the entire dataset is more efficient than sampling from medoids. This happens when the class' frequency is higher in the full dataset than its frequency in the cluster medoids. 

Overall, our results demonstrate the advantage of using the cluster medoids for collecting the initial annotated samples for a skewed dataset like Contract-NLI, which is a realistic use-case in the legal domain. It is noteworthy that the original version of LEDGAR dataset is also imbalanced, but as explained in Sec.~\ref{sec:datasets}, due to the drastically high number of classes, and for the sake of comparison with skewed datasets, only the most dominant categories are kept in this work. 

Thanks to the semantically meaningful and comparable sentence embeddings obtained after the knowledge distillation step, the cluster medoids well represent the entire dataset, and thus sampling among them drastically reduces the annotation effort without harming the performance. 
As a real life scenario, consider a company with hundreds of legal contracts aiming to classify their sentences into multiple categories, under a restricted budget. Reducing the annotation effort means lowering down the financial costs of annotation, which can be highly expensive in the legal domain (over $\$2$ million for annotating around $500$ contracts according to~\citet{Hendrycks2021CUADAE}). 


\begin{table*}[h]
\centering
\resizebox{\textwidth}{!}{
\begin{tabular}{clccccc}
\hline
\multirow{2}{*}{Dataset} & \multirow{2}{*}{Category} & \multicolumn{2}{c}{full dataset} & \multicolumn{2}{c}{medoids} & \multirow{2}{*}{gain(\%)} \\
& & median & $90^{th}\%$tile & median & $90^{th}\%$tile &  \\
\hline
\multirow{8}{*}{\small{\texttt{\rotatebox[origin=c]{90}{Contract-NLI}}}} 
& \small{\texttt{Inclusion of verbally conveyed information}} & 75.0 & 125.0 & 35.5 & 59.0 & 52.8\\
& \small{\texttt{No licensing}} & 64.0 & 108.0 & 68.5 & 109.1 & -1.0 \\
& \small{\texttt{No reverse engineering}} & 342.0 & 568.0 & 144.0 & 209.1 & 63.2\\
& \small{\texttt{Notice on compelled disclosure}} & 74.5 & 122.0 & 99.0 & 155.0 & -27.0\\
& \small{\texttt{Sharing with employees}} & 57.0 & 90.0 & 21.0 & 34.1 & 62.1 \\
& \small{\texttt{Sharing with third-parties}} & 54.0 & 92.1 & 21.0 & 35.0 & 62.0\\
& \small{\texttt{Survival of obligations}} & 64.0 & 106.0 & 36.0 & 57.0 & 46.2\\
& \small{\texttt{Return of confidential information}} & 116.0 & 189.0 & 61.0 & 99.0 & 47.6\\
\hline
\multirow{5}{*}{\small{\texttt{\rotatebox[origin=c]{90}{LEDGAR}}}} & \small{\texttt{Amendments}} & 23.0 & 37.1 & 21.0 & 33.0 & 10.8\\
& \small{\texttt{Counterparts}} & 26.0 & 42.0 & 34.0 & 54.1 & -28.8 \\
& \small{\texttt{Entire agreements}} & 26.0 & 42.0 & 33.0 & 55.0 & -30.9 \\
& \small{\texttt{Governing laws}} & 17.5 & 28.0 & 14.0 & 21.0 & 25.0\\
& \small{\texttt{Notices}} & 29.0 & 49.0 & 26.0 & 44.0 & 10.2\\
\hline
\end{tabular}}
\caption{\label{tab:exp_medoid}
Number of actions to acquire the initial labeled set for $8$ categories of Contract-NLI, and LEDGAR when sampling from the full dataset (standard approach), and sampling from the cluster medoids (our approach).}
\end{table*}



\begin{figure}[h!]
    \centering
    \includegraphics[width=0.5\textwidth]{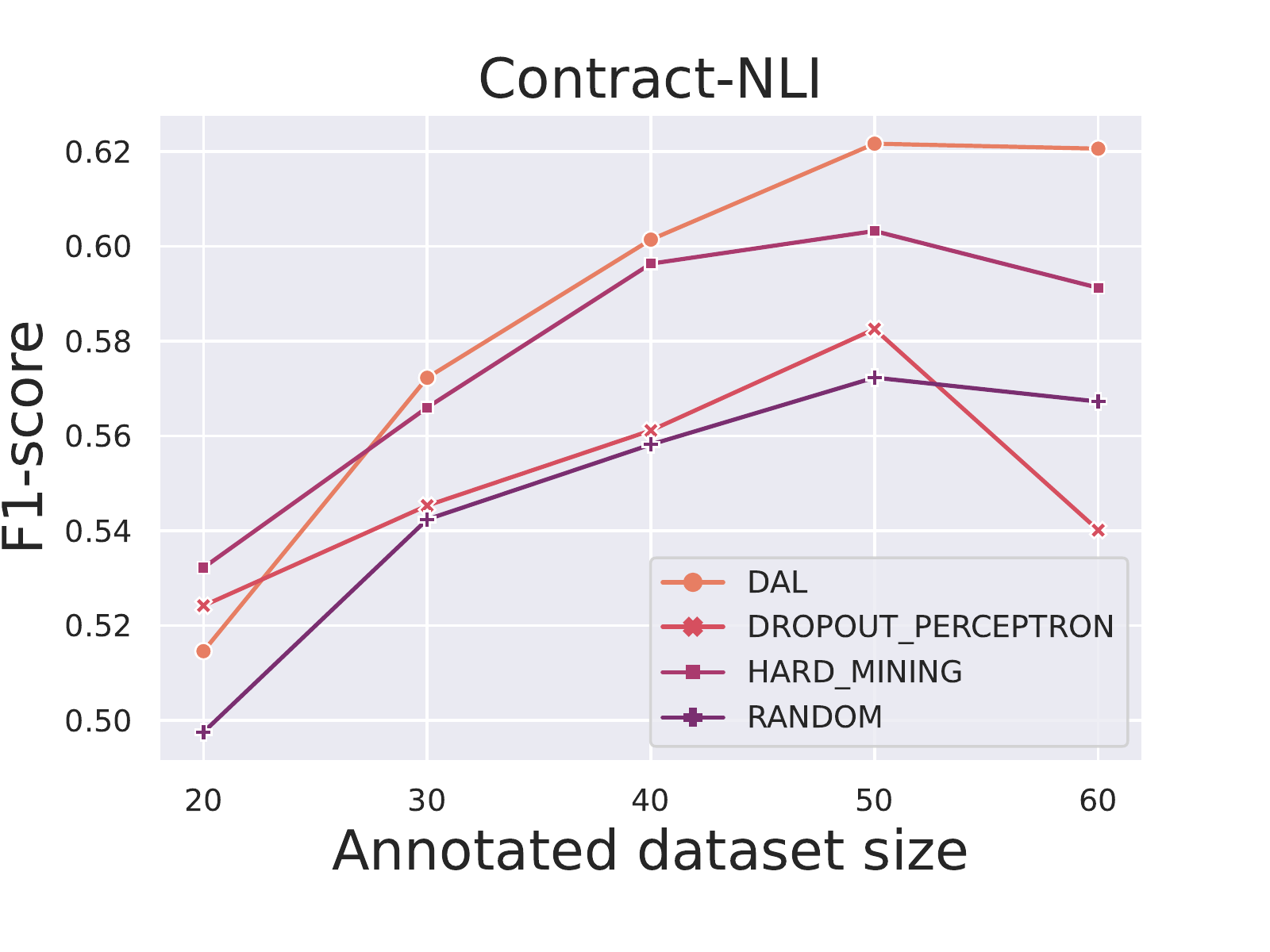}
    \includegraphics[width=0.5\textwidth]{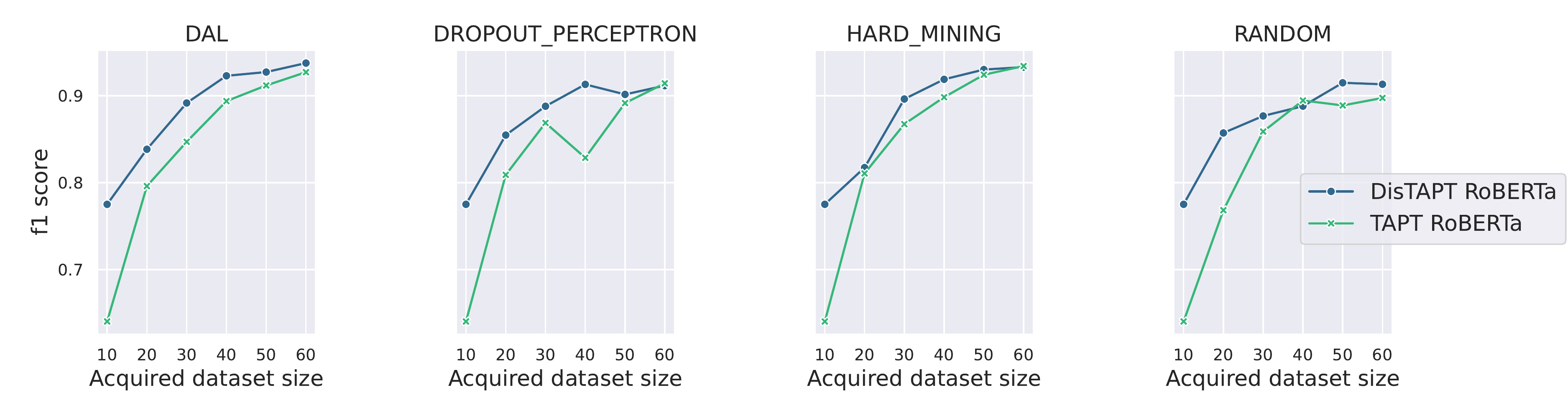}
    \caption{Comparison of four AL strategies when used with DisTAPT RoBERTa with IS.}
    \label{fig:all_ALs}
\end{figure}

\subsection{Effect of AL strategy}\label{sec:exp_AL_effect}

Finally, we evaluate the generalizability of our approach over the four AL strategies mentioned in Sec.~\ref{sec:setup}: DAL, Random, Hard-Mining, and Perceptron Dropout. As shown in Figure~\ref{fig:all_ALs}, DAL results in the best performance with at most $0.08$ higher F1-score than other strategies with $60$ labeled samples for Contract-NLI, and less than $0.04$ higher F1-score with $40$ annotated samples for LEDGAR. The small performance gap of these four AL methods in our pipeline indicates the generalizability of this approach to various AL strategies.

\section{Conclusion}

We propose an efficient active learning pipeline for legal text classification. Our approach leverages the available unlabeled data to adapt the pre-trained language model to the downstream task, and guide its embeddings to a semantically meaningful space before fine-tuning. We use model distillation to produce semantically comparable embeddings. A future work can study the effect of other approaches like BERT-Flow \citep{Li2020OnTS} and whitening \citep{Su2021WhiteningSR} on AL with this pipeline. Moreover, we design a simple strategy to efficiently acquire a labeled set of positive and negative samples for the initial iteration of active learning. 

Our experiments over Contract-NLI and LEDGAR benchmarks demonstrate the effectiveness of our approach compared to standard active learning strategies. Our results also show that our pipeline obtains very close performance to the fully-supervised approach with considerably less annotation cost. We test our methodology in the legal domain, and for four AL strategies, but we expect it to generalize to other strategies like ALPS and BADGE, and other specialized domains, like medicine. We leave this evaluation as a future work.

\section*{Limitations}

In this work, we have shown the importance of task-adaptation and knowledge distillation, and that we can leverage the available unlabeled data to perform efficient fine-tuning via active learning and obtain better performance. The price to pay for this performance gain is time and computational power. The time taken by task-adaptation and distillation scales with the size of unlabeled data. On the other hand, more unlabeled samples result in more effective adaptation to the downstream task. Therefore, the user of this approach needs to find the best trade-off given their data, annotation budget, time and computational power. For, LEDGAR, the larger dataset used in this work, we performed the adaptation and distillation steps in $4$ and $1$ hour(s) respectively, using a single Nvidia GeForce GTX TITAN X GPU.

Moreover, we showed that by clustering the sentence embeddings produced by DisTAPT RoBERTa, the initial labeled set can be acquired more efficiently. Nevertheless, this approach inherits the limitations of clustering. Namely, the time complexity of clustering the embeddings scales with the data, and the number of clusters should be empirically chosen. In our experiments we spent $10$ minutes to cluster the $44,249$ samples belonging to LEDGAR dataset into $442$ groups.

\section*{Ethics Statement}
Industries have hundreds of contracts with tens of thousands of sentences that belong to various topics. Labeling all of these samples is a highly expensive and time-consuming process. In this work, we aim to reduce the resources spent on this task by leveraging recent advances in natural language processing, while keeping the human expert in the loop. The goal is to reduce the human effort in annotation so that the legal experts' time and knowledge can be used in another task at which humans are better than machines. 





\bibliographystyle{acl_natbib}

\appendix

\section{Appendix}\label{sec:appendix}

\subsection{Dataset Distributions}\label{sec:app_class_dist}

We provide the details of class distributions for Contract-NLI and LEDGAR benchmarks in Table~\ref{tab:app_freq}. As shown in this table, LEDGAR contains considerably larger categories compared to Contract-NLI and is more balanced.

\begin{table*}[h]
\centering
\begin{tabular}{clccc}
\hline
Dataset & Category & Train Size & Dev Size & Test Size\\
\hline
\multirow{17}{*}{\texttt{Contract-NLI}} & \small{\texttt{Confidentiality of Agreement}} & 161 & 29 & 46\\
& \small{\texttt{Explicit identification}} & 203 & 29 & 60\\
& \small{\texttt{Inclusion of verbally conveyed information}} & 274 & 45 & 76\\
& \small{\texttt{Limited use}} & 371 & 53 & 110\\
& \small{\texttt{No licensing}} & 327 & 39 & 86\\
& \small{\texttt{No reverse engineering}} & 60 & 8 & 13\\
& \small{\texttt{No solicitation}} & 93 & 11 & 28\\
& \small{\texttt{None-inclusion of non-technical information}} & 332 & 50 & 94\\
& \small{\texttt{Notice on compelled disclosure}} & 276 & 45 & 77\\
& \small{\texttt{Permissible acquirement of similar information}} & 311 & 47 & 96\\
& \small{\texttt{Permissible copy}} & 167 & 17 & 49\\
& \small{\texttt{Permissible development of similar information}} & 263 & 40 & 73\\
& \small{\texttt{Permissible post-agreement possession}} & 312 & 25 & 63\\
& \small{\texttt{Return of confidential information}} & 182 & 24 & 38\\
& \small{\texttt{Sharing with employees}} & 358 & 56 & 94\\
& \small{\texttt{Sharing with third-parties}} & 370 & 53 & 102\\
& \small{\texttt{Survival of obligations}} & 311 & 43 & 83\\
\hline

\multirow{5}{*}{\texttt{LEDGAR}} & \small{\texttt{Amendments}} & 9,132 & 1,515 & 2,615 \\
& \small{\texttt{Counterparts}} & 8,033 & 1,312 & 2,363 \\
& \small{\texttt{Entire agreements}} & 8,094 & 1,361 & 2,370 \\
& \small{\texttt{Governing laws}} & 11,926 & 1,997 & 3,454 \\
& \small{\texttt{Notices}} & 7,064 & 1,190 & 2,105 \\
\hline
\end{tabular}
\caption{\label{tab:app_freq}
Category frequency for Contract-NLI adapted to classification task, and LEDGAR benchmarks.}
\end{table*}

\subsection{Effective Fine-Tuning}\label{sec:app_tabs}

\begin{figure}[t]
    \centering
    \includegraphics[width=0.5\textwidth]{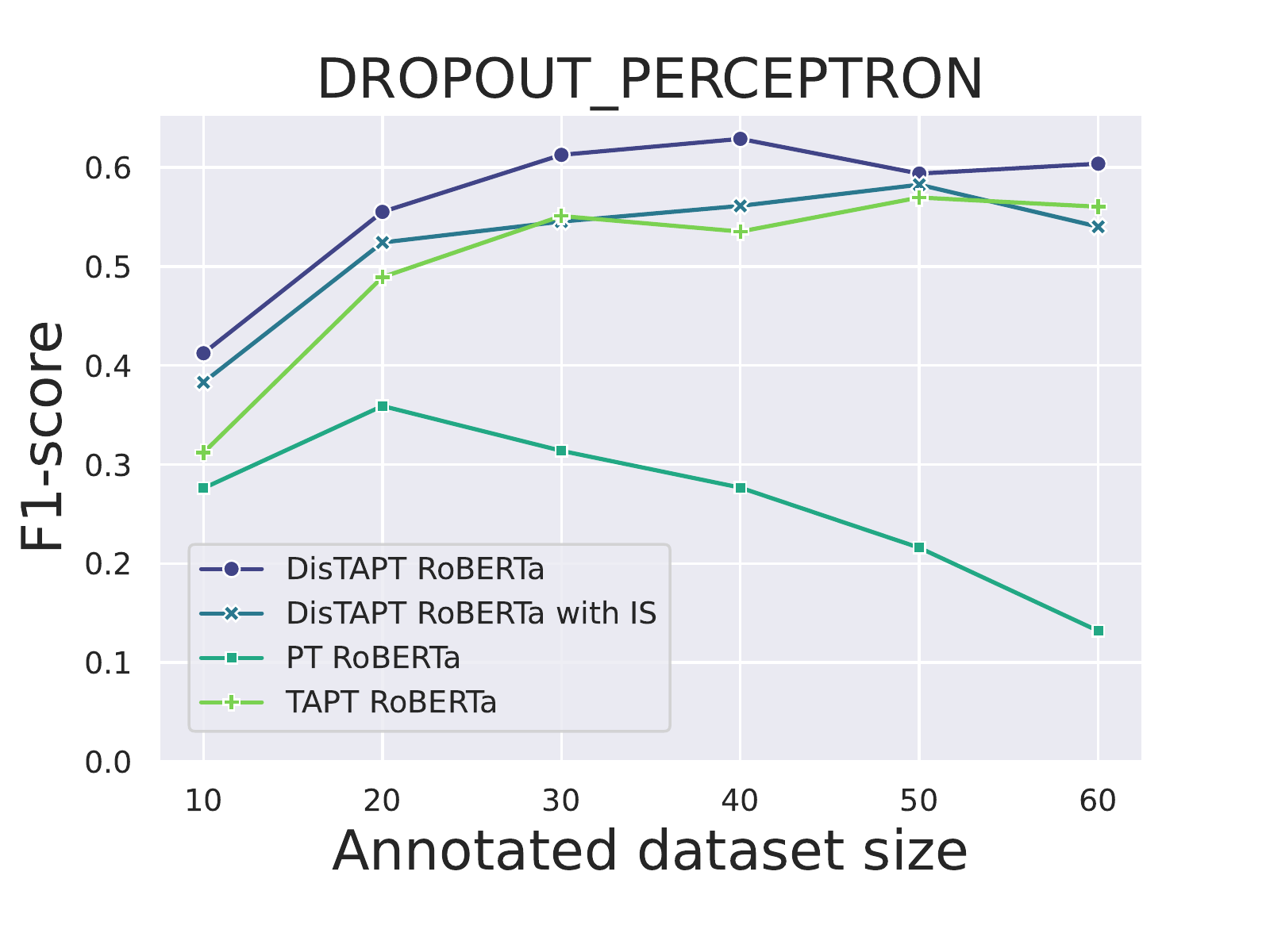}
    \includegraphics[width=0.5\textwidth]{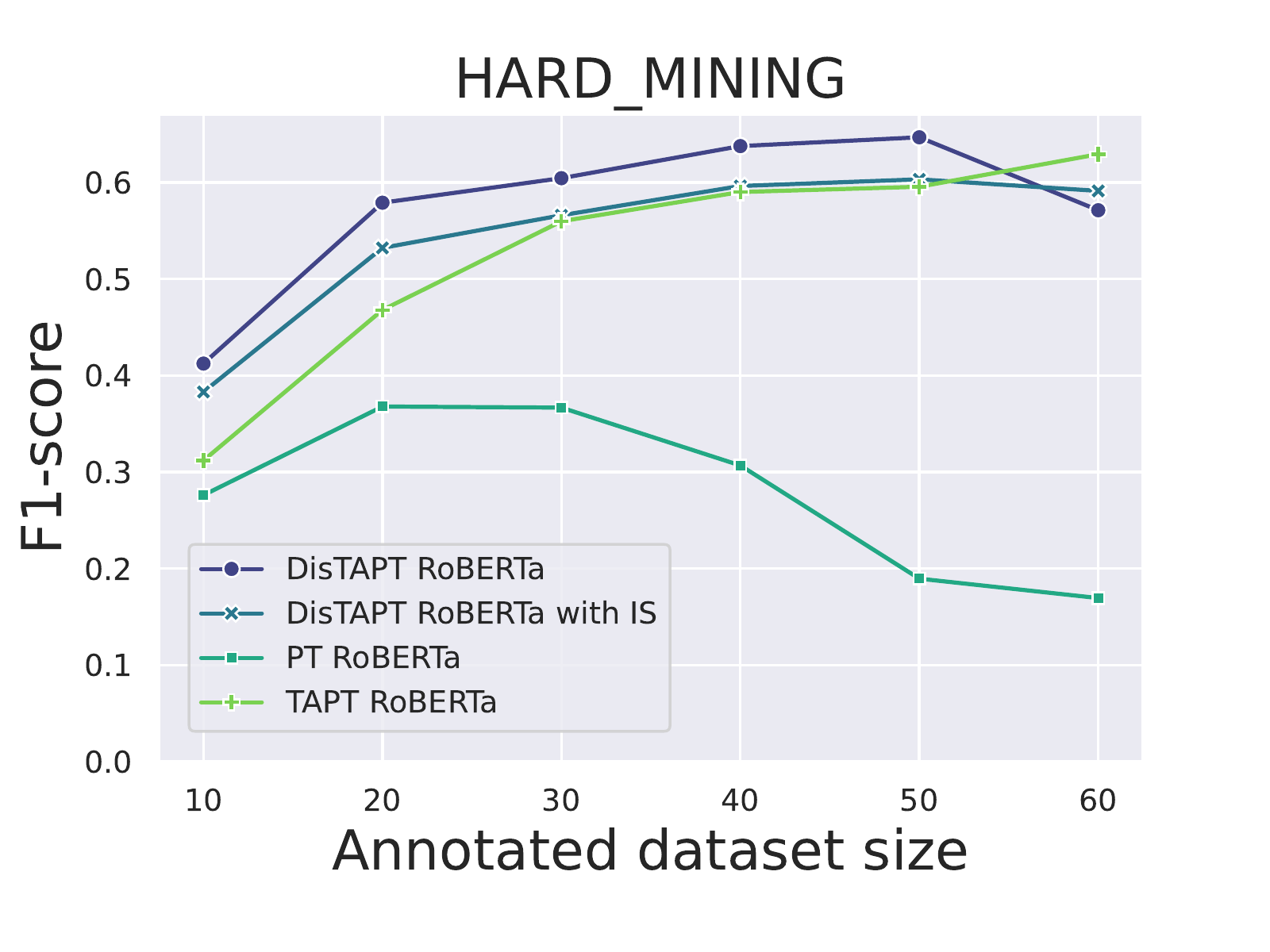}
    \includegraphics[width=0.5\textwidth]{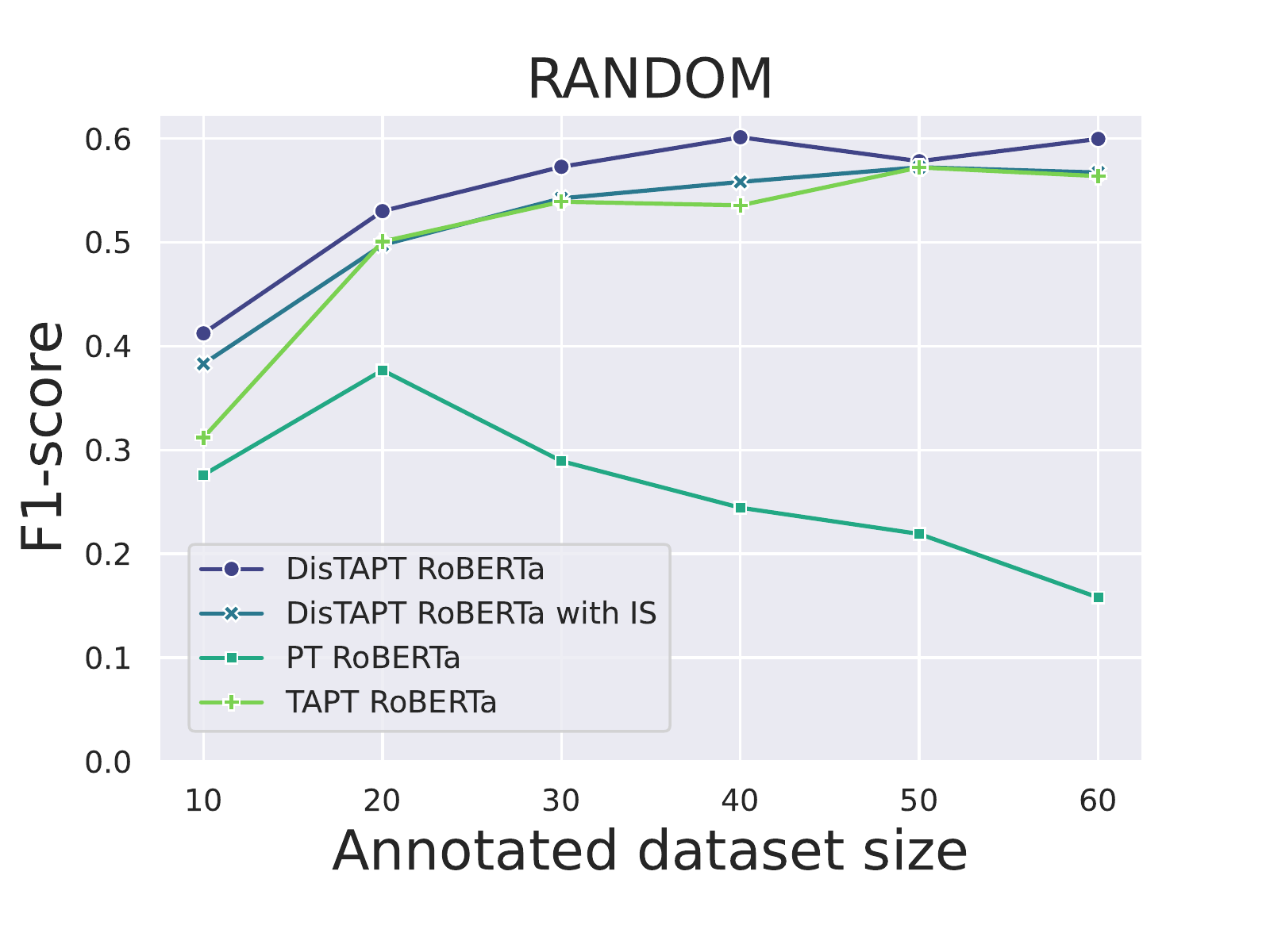}
    \caption{Test F1-score for \textbf{Contract-NLI} during AL iterations. The F1-score for the fully supervised fine-tuning is $0.6990$.}
    \label{fig:app_contract_nli}
\end{figure}

\begin{figure}[t]
    \centering
    \includegraphics[width=0.5\textwidth]{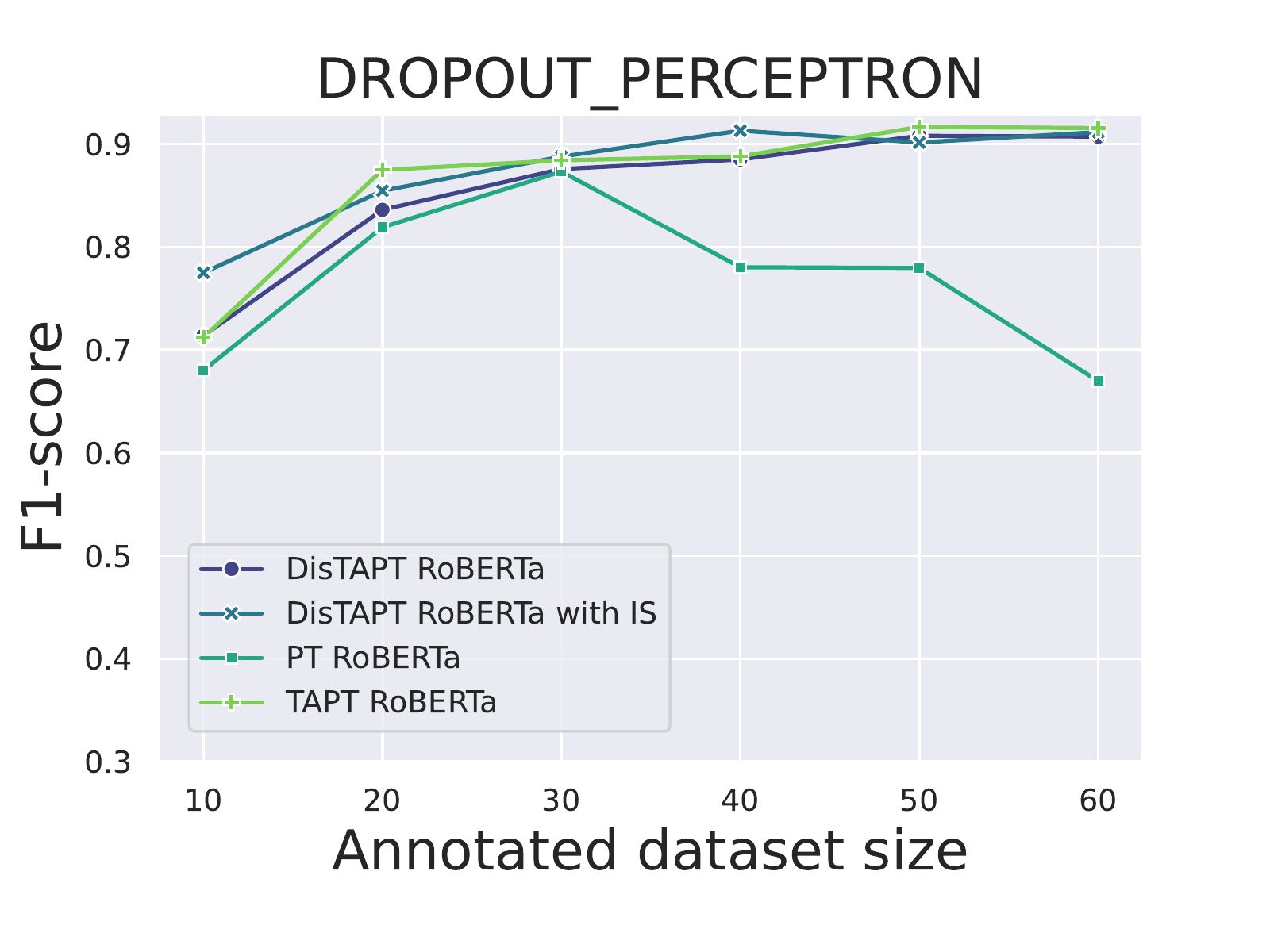}
    \includegraphics[width=0.5\textwidth]{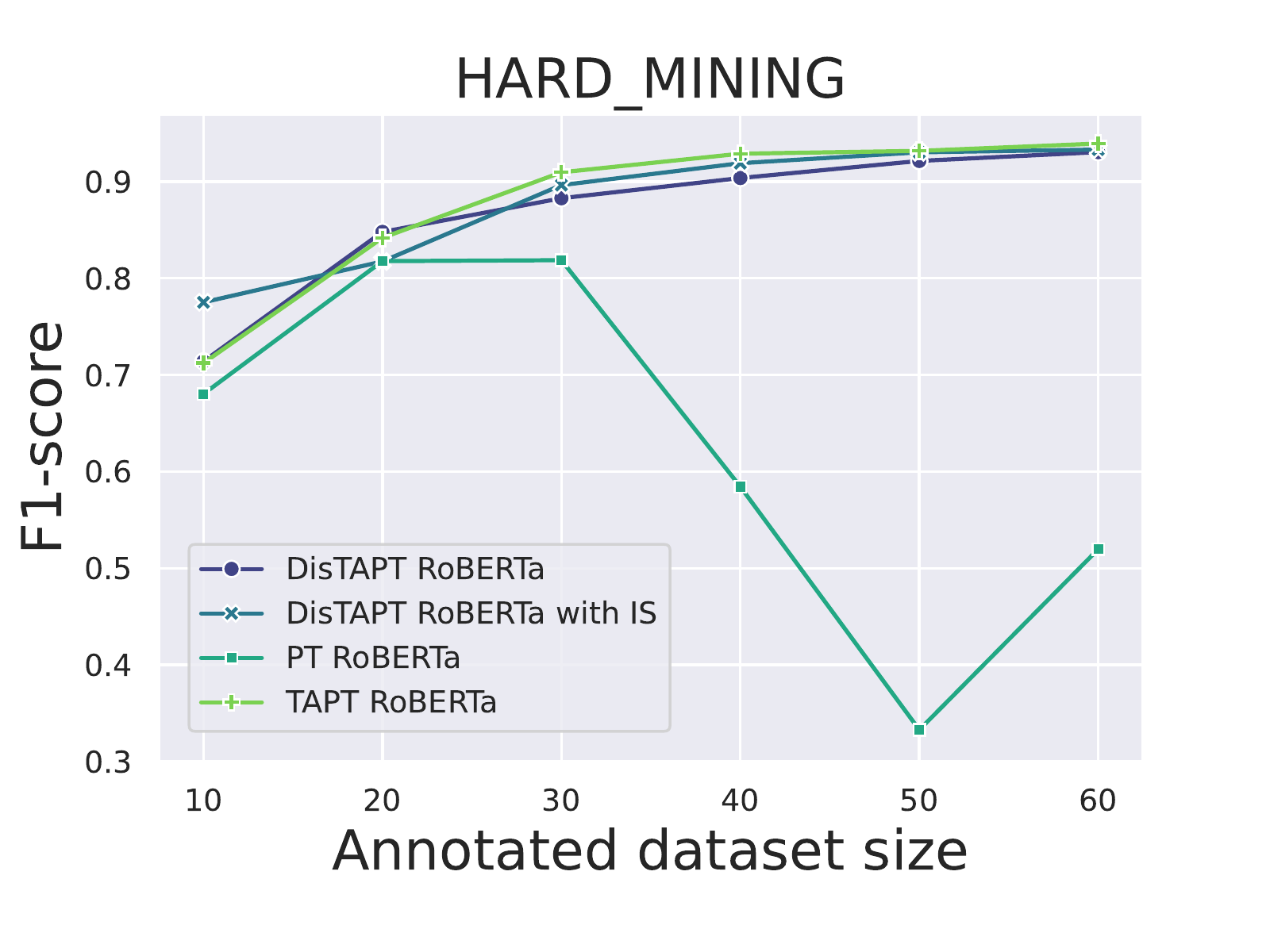}
    \includegraphics[width=0.5\textwidth]{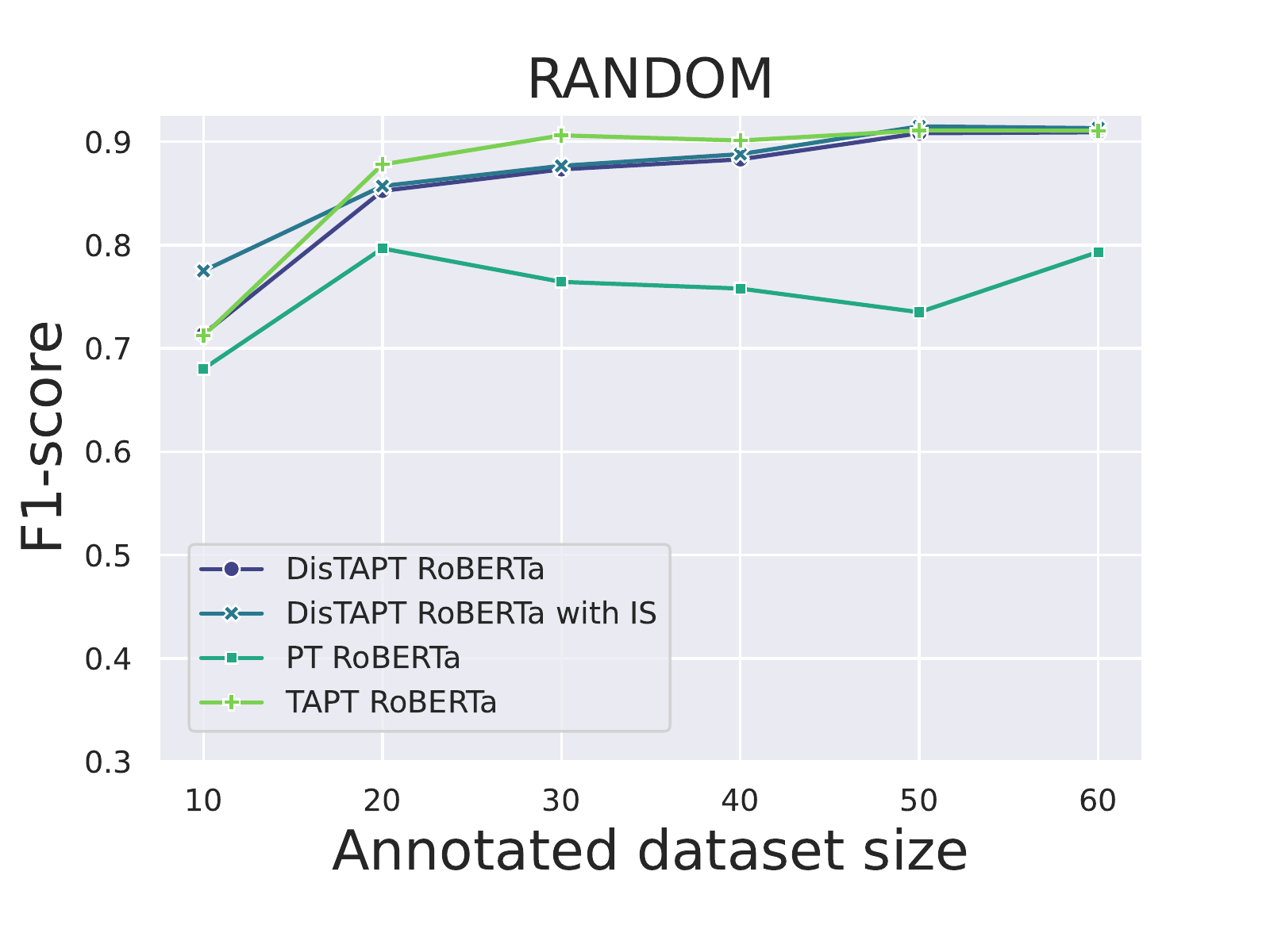}
    \caption{Test F1-score for \textbf{LEDGAR} during AL iterations. The F1-score for the fully supervised fine-tuning is $0.9538$.}
    \label{fig:app_ledgar}
\end{figure}

Here we present the results of standard active learning and our approach for four AL strategies discussed in Sec.~\ref{sec:setup} including Random, Hard-Mining, and Perceptron Dropout. As before, we report the average F1-score over three runs. Figure~\ref{fig:app_contract_nli} corresponds to Contract-NLI and Figure~\ref{fig:app_ledgar} illustrates the results for the LEDGAR dataset.

\subsection{Effect of Knowledge Distillation}\label{sec:app_knowldge_distill}

\begin{figure}[t]
    \centering
    \includegraphics[width=0.4\textwidth]{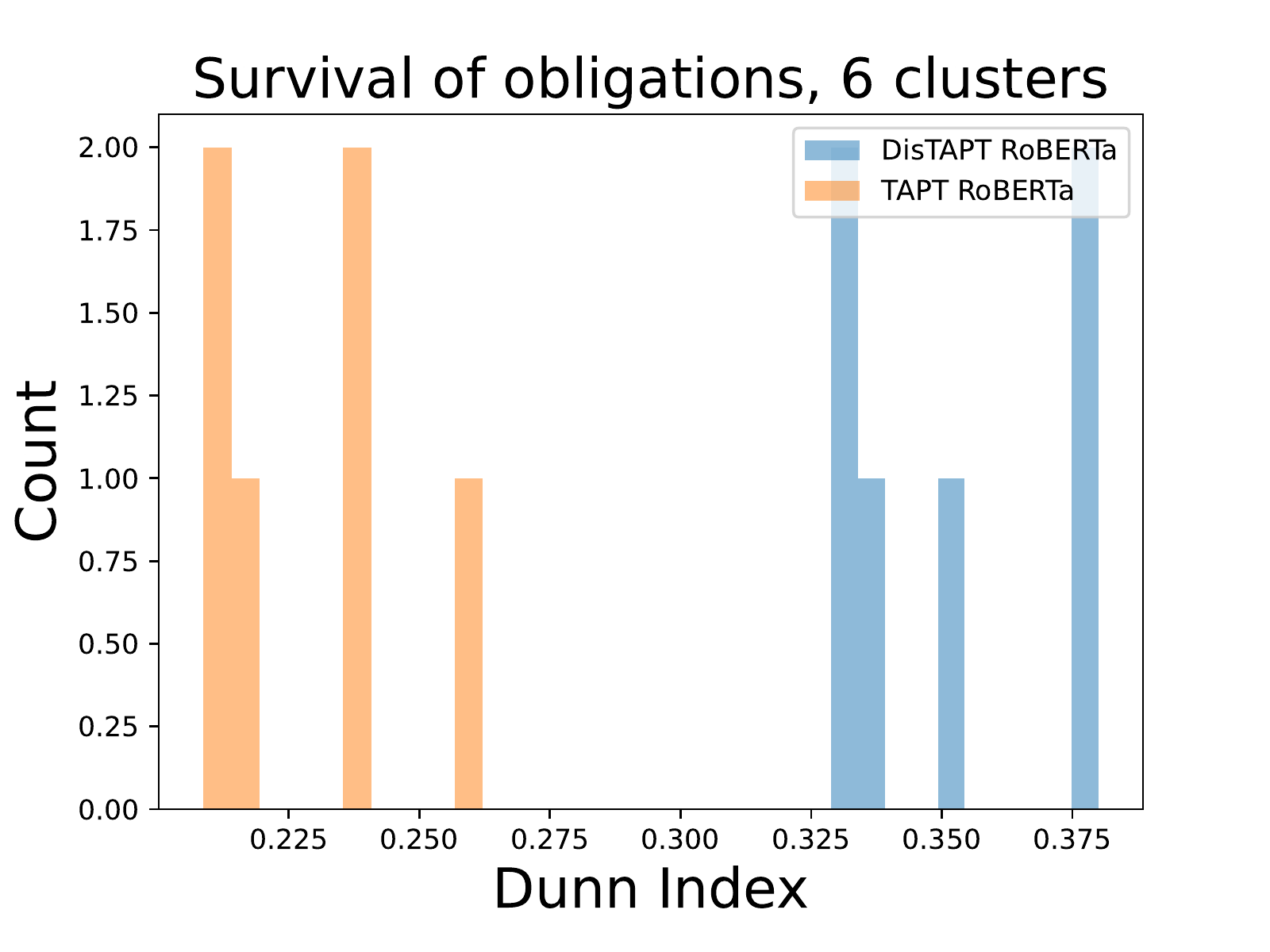}
    \includegraphics[width=0.4\textwidth]{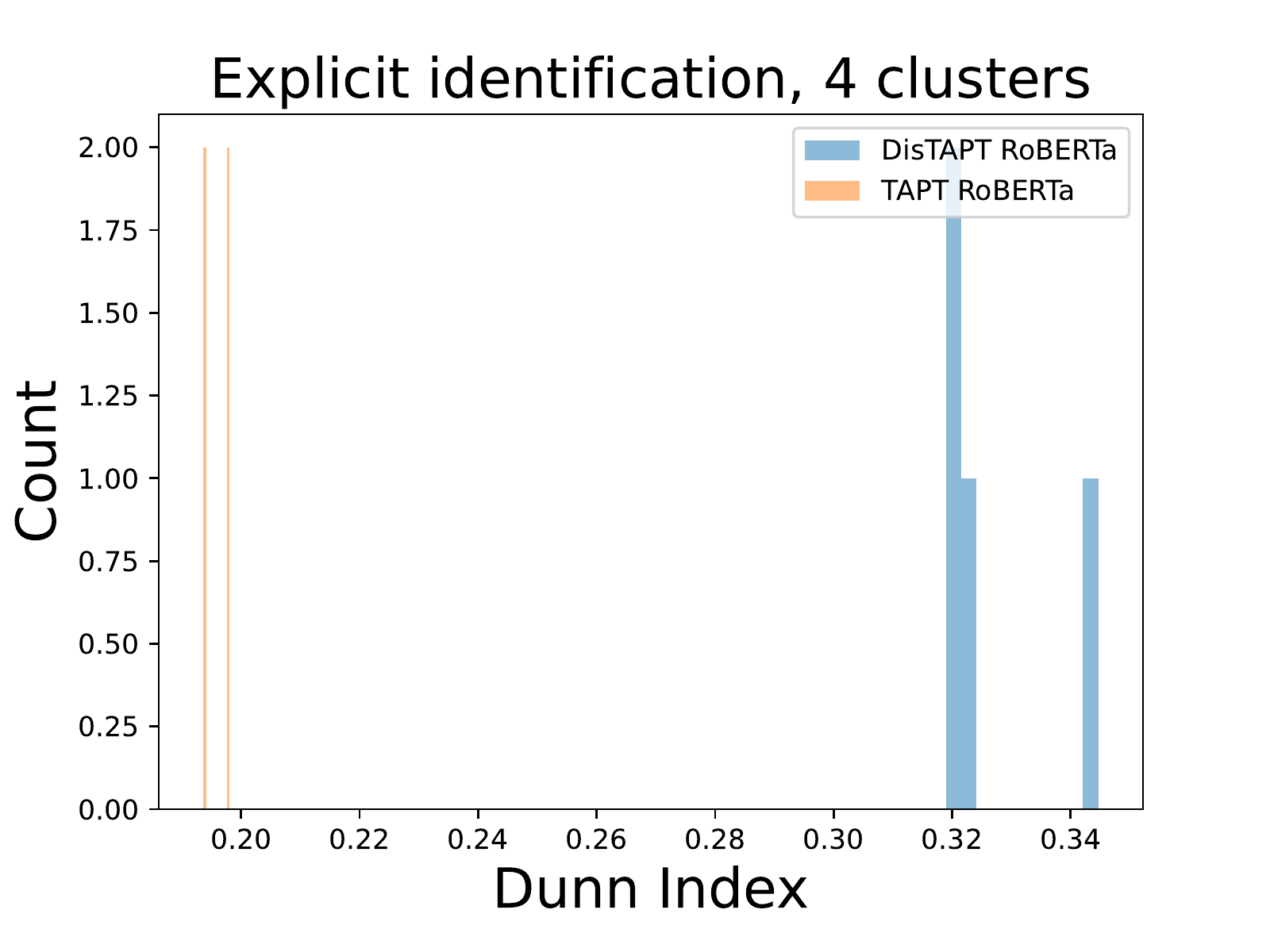}
    \includegraphics[width=0.4\textwidth]{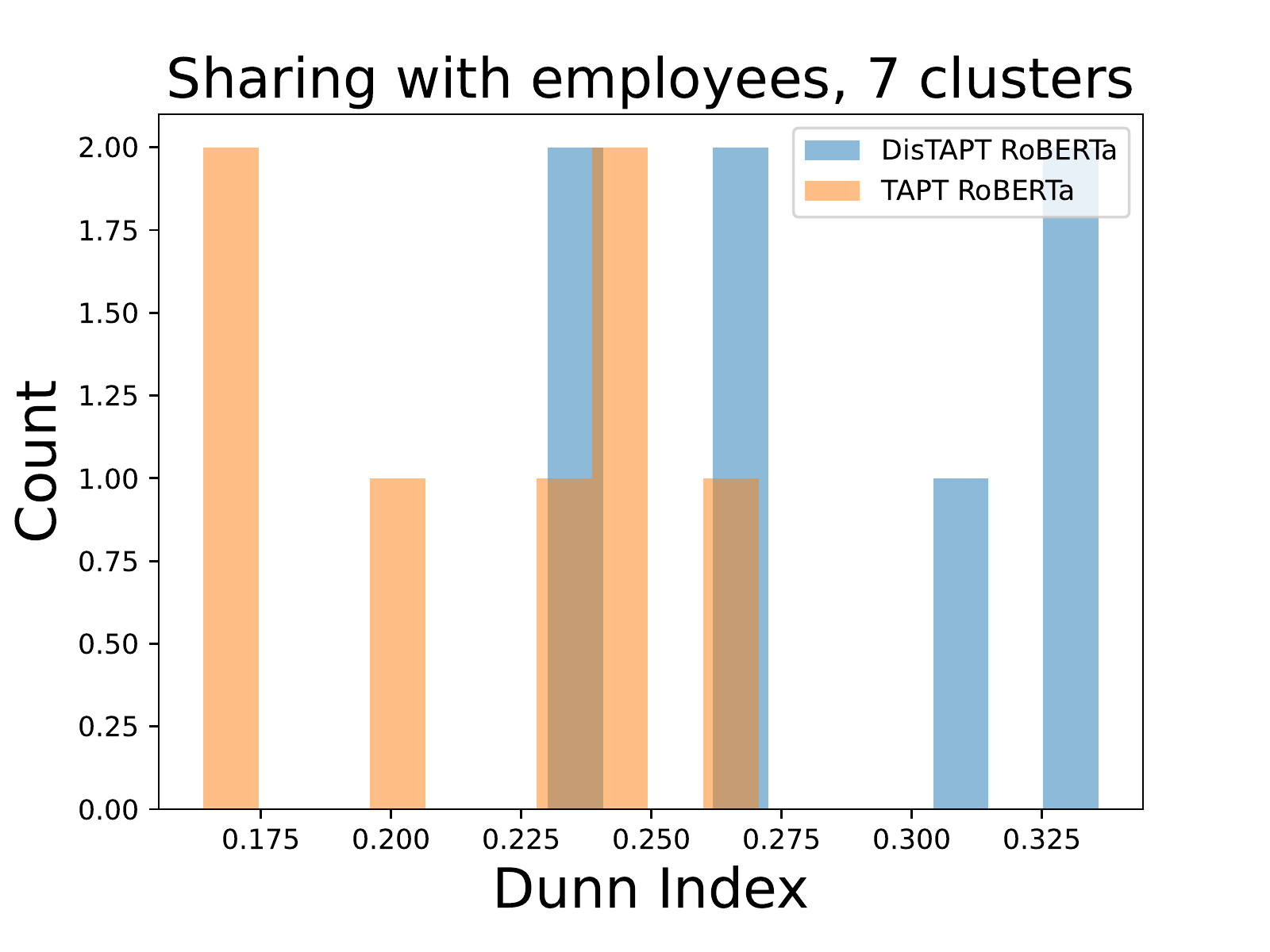}
    \includegraphics[width=0.4\textwidth]{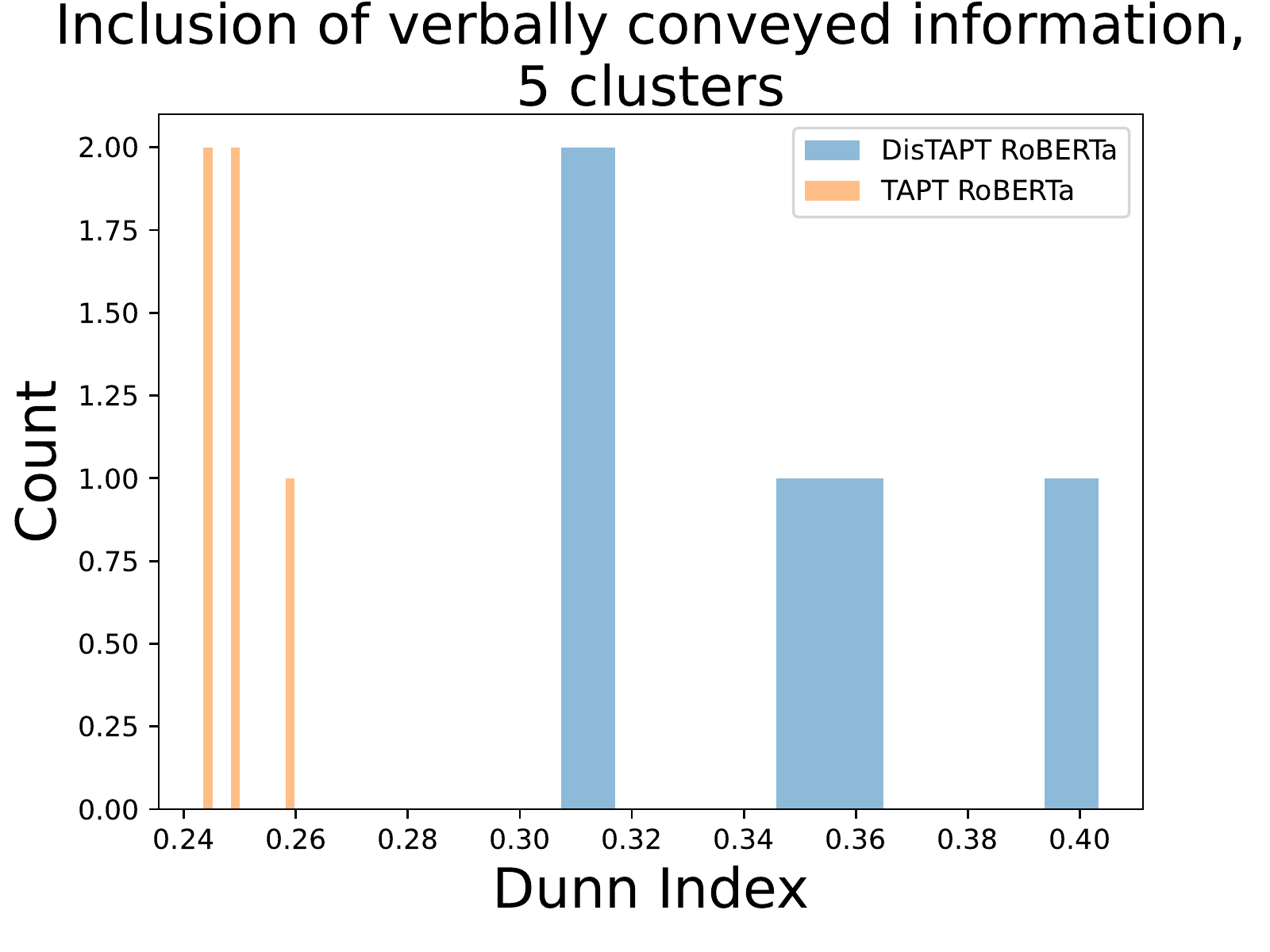}
    \includegraphics[width=0.4\textwidth]{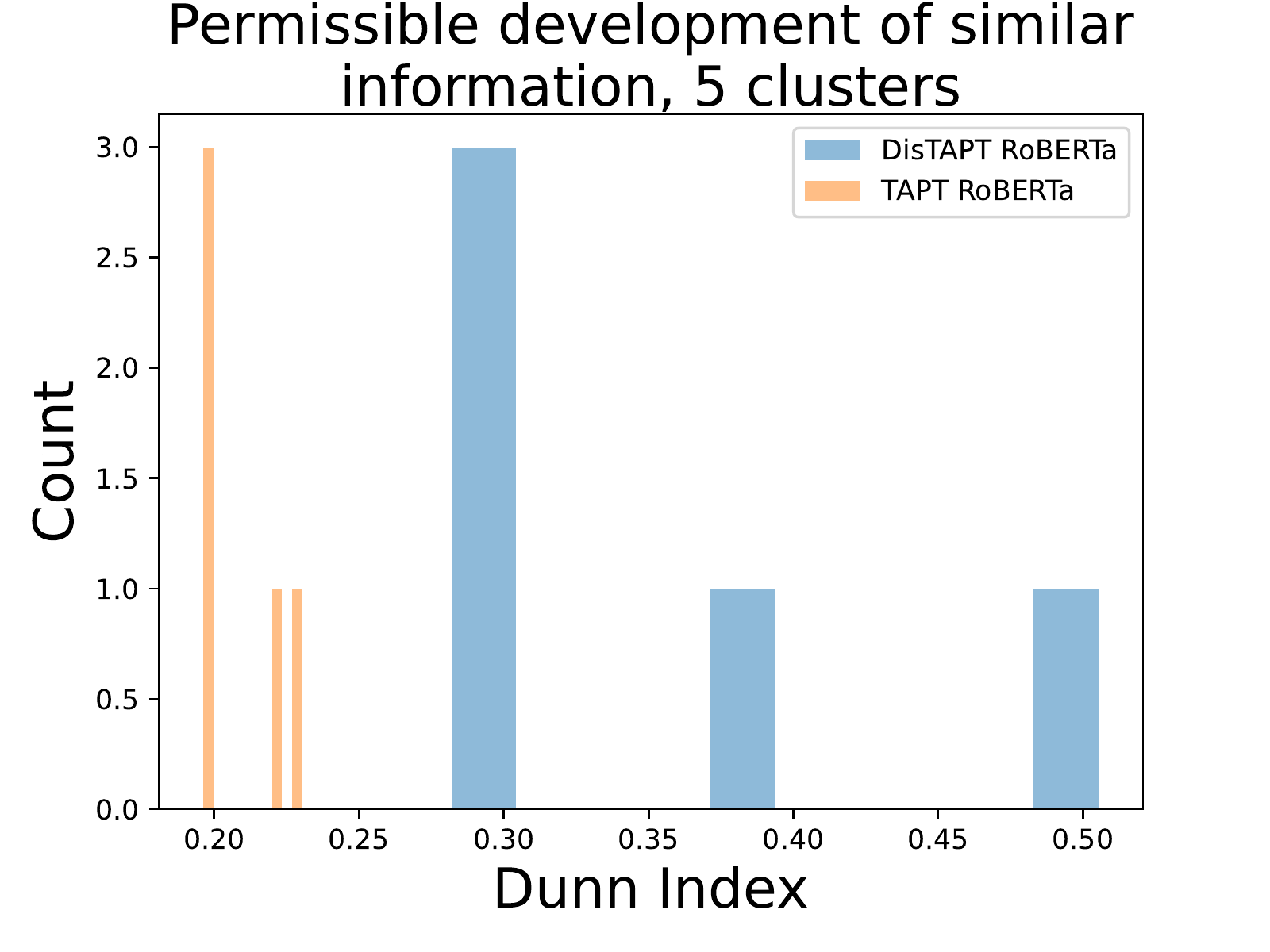}
    \caption{Comparison of the Dunn Index distribution before (TAPT RoBERTa) and after knowledge distillation (DisTAPT RoBERTa) for \textbf{Contract-NLI} dataset.}
    \label{fig:app_contract-nli-distill-effect-clusters}
\end{figure}

\begin{figure}[t]
    \centering
    \includegraphics[width=0.4\textwidth]{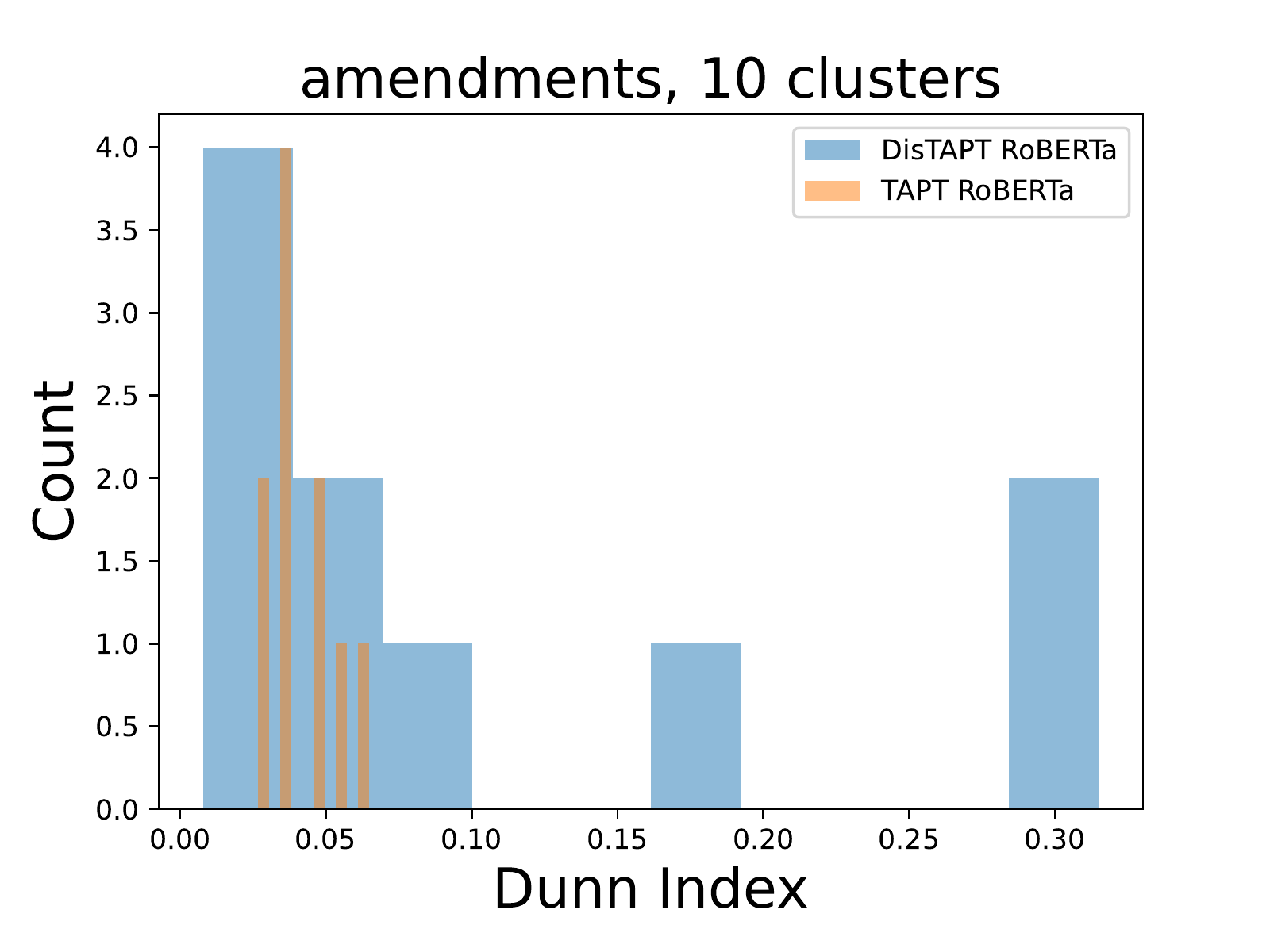}
    \includegraphics[width=0.4\textwidth]{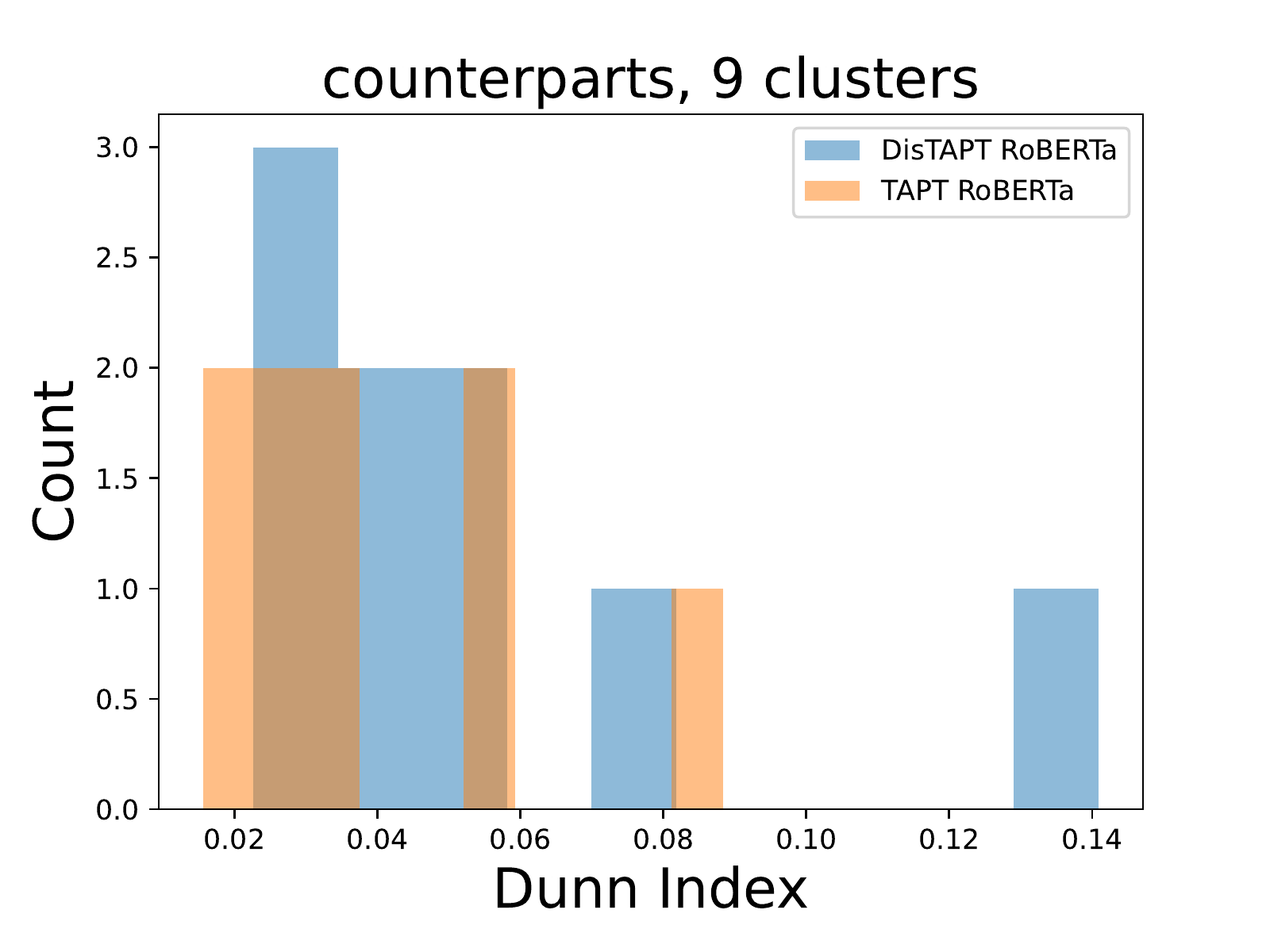}
    \includegraphics[width=0.4\textwidth]{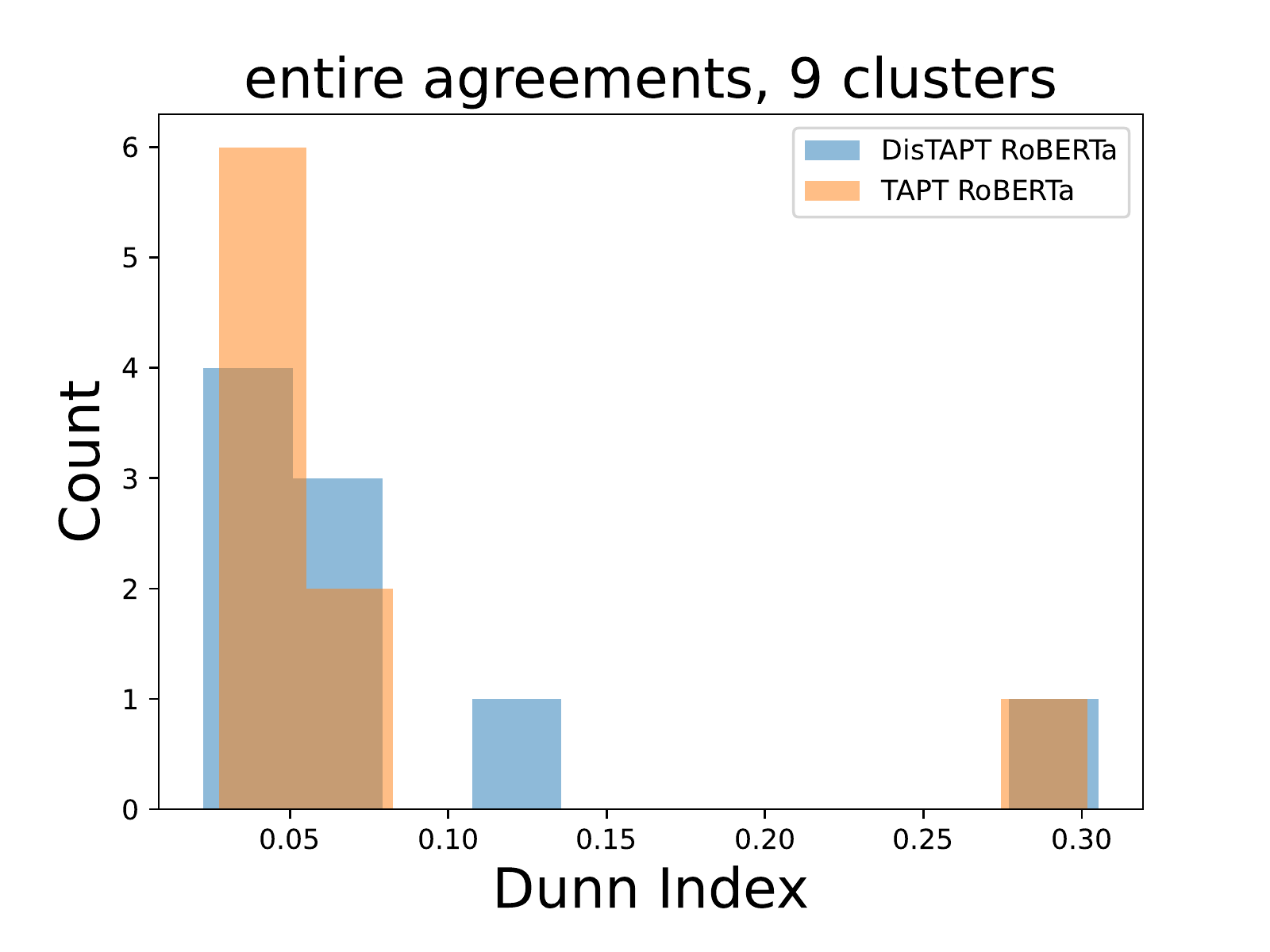}
    \includegraphics[width=0.4\textwidth]{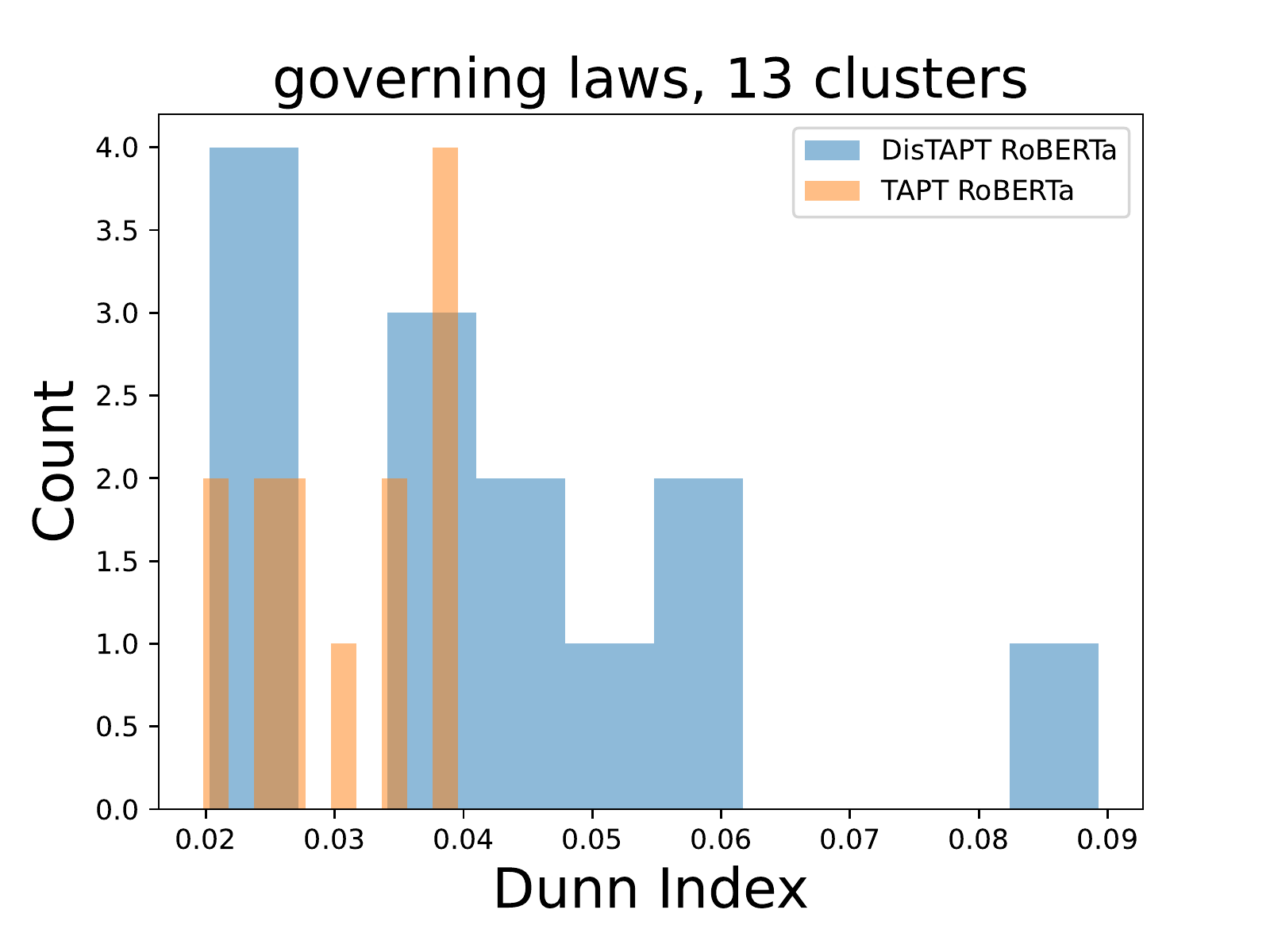}
    \includegraphics[width=0.4\textwidth]{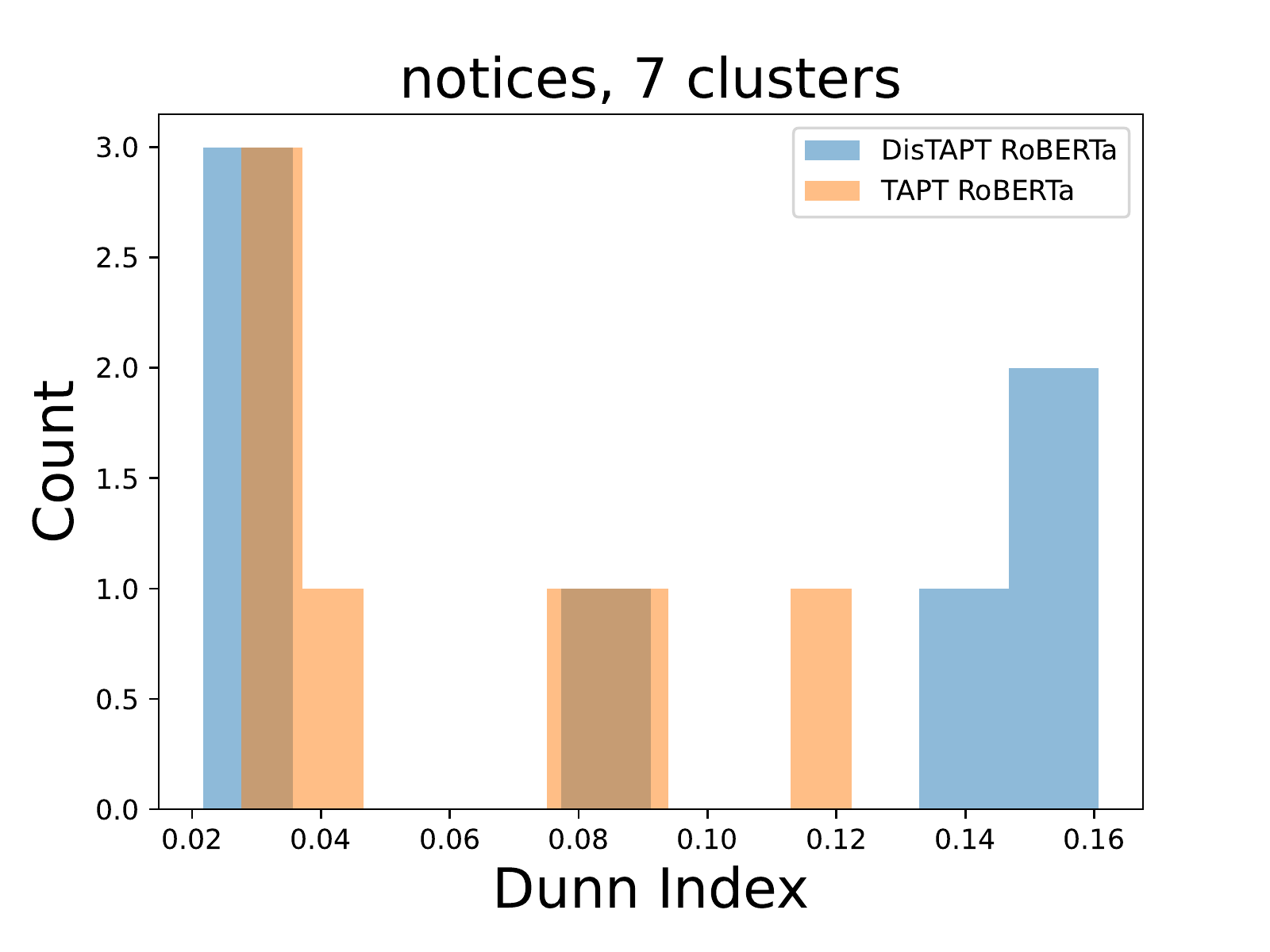}
    \caption{Comparison of the Dunn Index distribution before (TAPT RoBERTa) and after knowledge distillation (DisTAPT RoBERTa) for \textbf{LEDGAR} dataset.}
    \label{fig:app_ledgar-distill-effect-clusters}
\end{figure}

Figures~\ref{fig:app_contract-nli-distill-effect-clusters} and~\ref{fig:app_ledgar-distill-effect-clusters} illustrate the comparison of the Dunn Index distribution that were not presented in the main paper. 

\subsection{Efficiency of Initial Sampling with Medoids}\label{sec:app_medoid}

In Table~\ref{tab:app_medoid} we provide the median and $90^{th}$ percentile of number of actions performed to collect the initial labeled set, for the standard sampling approach, and our proposed strategy using cluster medoids, for nine categories of Contract-NLI that were not included in Table~\ref{tab:exp_medoid} in Sec.~\ref{sec:exp_initial-sampling}. It is observed that, for most categories, there is a considerable reduction in the number of actions performed to acquire the annotated data for the initial AL iteration. 

\begin{table*}[h]
\centering
\begin{tabular}{lccccc}
\hline
\multirow{2}{*}{Category} &  \multicolumn{2}{c}{full dataset} & \multicolumn{2}{c}{medoids} & \multirow{2}{*}{gain($\%$)}\\
 & median & $90^{th}\%$tile & median & $90^{th}\%$tile \\
\hline
\small{\texttt{Confidentiality of Agreement}} & 125.0 & 215.1 & 120.0 & 178.2 & 17.1\\
\small{\texttt{Explicit identification}} & 100.0 & 161.1 & 48.0 & 77.0 & 52.2\\
\small{\texttt{Limited use}} & 56.0 & 90.1 & 37.0 & 58.0 & 35.6\\
\small{\texttt{No solicitation}} & 227.0 & 383.0 & 178.0 & 261.0 & 31.8\\
\small{\texttt{None-inclusion of non-technical information}} & 61.0 & 101.1 & 39.0 & 64.0 & 36.7\\
\small{\texttt{Permissible acquirement of similar information}} & 65.0 & 107.0 & 91.0 & 145.0 & -35.5\\
\small{\texttt{Permissible copy}} & 121.0 & 197.0 & 68.0 & 108.0 & 45.2\\
\small{\texttt{Permissible development of similar information}} & 77.0 & 129.1 & 82.0 & 129.0 & 0.1\\
\small{\texttt{Permissible post-agreement possession}} & 66.0 & 108.1 & 41.0 & 66.0 & 38.9\\
\hline
\end{tabular}
\caption{\label{tab:app_medoid}
Number of actions to acquire the initial labeled set for $9$ categories of Contract-NLI when sampling from the full dataset (standard approach), and sampling from the cluster medoids.}
\end{table*}

\end{document}